# What's in an Attribute?
# Consequences for the Least Common Subsumer

**Ralf Küsters**                                    KUESTERS@TI.INFORMATIK.UNI-KIEL.DE
*Institut für Informatik und Praktische Mathematik*
*Christian-Albrechts-Universität zu Kiel*
*24098 Kiel*
*Germany*

**Alex Borgida**                                         BORGIDA@CS.RUTGERS.EDU
*Department of Computer Science*
*Rutgers University*
*Piscataway, NJ 08855*
*USA*

## Abstract

Functional relationships between objects, called "attributes", are of considerable importance in knowledge representation languages, including Description Logics (DLs). A study of the literature indicates that papers have made, often implicitly, different assumptions about the nature of attributes: whether they are always required to have a value, or whether they can be partial functions. The work presented here is the first explicit study of this difference for subclasses of the CLASSIC DL, involving the same-as concept constructor. It is shown that although determining subsumption between concept descriptions has the same complexity (though requiring different algorithms), the story is different in the case of determining the least common subsumer (lcs). For attributes interpreted as partial functions, the lcs exists and can be computed relatively easily; even in this case our results correct and extend three previous papers about the lcs of DLs. In the case where attributes must have a value, the lcs may not exist, and even if it exists it may be of exponential size. Interestingly, it is possible to decide in polynomial time if the lcs exists.

## 1. Introduction

Knowledge representation systems based on Description Logics (DLs) have been the subject of continued attention in Artificial Intelligence, both as a subject of theoretical studies (Borgida, 1994; Baader, 1996; Baader & Sattler, 2000; Giacomo & Lenzerini, 1996; Calvanese, Giacomo, & Lenzerini, 1999b) and in applications (Artale, Franconi, Guarino, & Pazzi, 1996; Brachman, McGuinness, Patel-Schneider, & Borgida, 1999; McGuinness & Patel-Schneider, 1998). More impressively, DLs have found applications in other areas involving information processing, such as databases (Borgida, 1995; Calvanese, Lenzerini, & Nardi, 1999), semi-structured data (Calvanese, Giacomo, & Lenzerini, 1998, 1999a), information integration (Calvanese, Giacomo, Lenzerini, Nardi, & Rosati, 1998; Borgida & Küsters, 2000), as well as more general problems such as configuration (McGuinness & Wright, 1998) and software engineering (Borgida & Devanbu, 1999; Devanbu & Jones, 1997). In fact, wherever the ubiquitous term "ontology" is used these days (e.g., for pro-viding the semantics of web/XML documents), DLs are prime contenders because of their clear semantics and well-studied computational properties.

In Description Logics, one takes an object-centered view, where the world is modeled as individuals, connected by binary relationships (here called *roles*), and grouped into classes (called *concepts*). For those more familiar with Predicate Logic, objects correspond to constants, roles to binary predicates, and concepts to unary predicates. In every DL system, the concepts of the application domain are described by *concept descriptions* that are built from atomic concepts and roles using the "constructors" provided by the DL language. For example, consider a situation where we want a concept describing individual cars that have had frequent (at least 10) repairs, and also record the fact that for cars, their model is the same as their manufacturer's model. Concepts can be thought of as being built up from (possibly nested) simpler noun-phrases, so the above concept, called Lemon in the sequel, might be captured as the conjunction of

(objects that are Cars)
(things **all** of whose model values are in concept Model)
(things **all** of whose madeBy values are in concept Manufacturer)
(things whose model value is **the same as** the model **of the** madeBy attribute)
(things with **at least** 10 repairs values)
(things **all** of whose repairs values are RepairReport).

Using the syntax of the CLASSIC language, we can abbreviate the above, while emphasizing the term-like nature of descriptions and the constructors used in each:

(**and** Car
    (**all** model Model)
    (**all** madeBy Manufacturer)
    (**same-as** (model) (madeBy ∘ model))
    (**at-least** 10 repairs)
    (**all** repairs RepairReport))

So, for example, the concept term (**at-least** $n$ **p**) has constructor **at-least**, and denotes objects which are related by the relationship **p** to at least $n$ other objects; in turn, (**all p C**) has as instances exactly those objects which are related by **p** only to instances of **C**.

Finally, we present the same concept in a mathematical notation which is more succinct and preferred in formal work on DLs:

Lemon :=   Car ⊓
        ∀model.Model ⊓
        ∀madeBy.Manufacturer ⊓
        madeBy ↓ (model ∘ madeBy) ⊓
        ≥ 10 repairs ⊓
        ∀repairs.RepairReport

Unlike preceding formalisms, such as semantic networks and frames (Quillian, 1968; Minsky, 1975), DLs are equipped with a formal semantics, which can be given by a translation into first-order predicate logic (Borgida, 1994), for example. Moreover, DL systems provide their users with various inference capabilities that allow them to deduce implicit knowledge from the explicitly represented knowledge. For instance, the subsumption algorithms allow one to determine subconcept-superconcept relationships: $C$ is *subsumed by* $D$ ($C \sqsubseteq D$) if and only if all instances of $C$ are also instances of $D$, i.e., the first description is always interpreted as a subset of the second description. For example, the concept Car obviously subsumes the concept description Lemon, while (**at-least** 10 repairs) is subsumed by (**at-least** 8 repairs).

The traditional inference problems for DL systems, such as subsumption, inconsistency detection, membership checking, are by now well-investigated. Algorithms and detailed complexity results for realizing such inferences are available for a variety of DLs of differing expressive power — see, e.g., (Baader & Sattler, 2000) for an overview.

## 1.1 Least Common Subsumer

The *least common subsumer (lcs)* of concepts is the most specific concept description subsuming the given concepts. Finding the lcs was first introduced as a new inference problem for DLs by Cohen, Borgida, and Hirsh (1992). One motivation for considering the lcs is to use it as an alternative to disjunction. The idea is to replace disjunctions like $C_1 \sqcup \cdots \sqcup C_n$ by the lcs of $C_1, \ldots, C_n$. Borgida and Etherington (1989) call this operation *knowledge-base vivification*. Although, in general, the lcs is not equivalent to the corresponding disjunction, it is the best approximation of the disjunctive concept within the available language. Using such an approximation is motivated by the fact that, in many cases, adding disjunction would increase the complexity of reasoning.[1]

As proposed by Baader et al. (Baader & Küsters, 1998; Baader, Küsters, & Molitor, 1999), the lcs operation can be used to support the "bottom-up" construction of DL knowledge bases, where, roughly speaking, starting from "typical" examples an lcs algorithm is used to compute a concept description that (i) contains all these examples, and (ii) is the most specific description satisfying property (i). Baader and Küsters have presented such an algorithm for cyclic $\mathcal{ALN}$-concept descriptions; $\mathcal{ALN}$ is a relatively simple language allowing for concept conjunction, primitive negation, value restrictions, and number restrictions. Also, Baader et al. (1999) have proposed an lcs algorithm for a DL allowing existential restrictions instead of number restrictions.

Originally, the lcs was introduced as an operation in the context of inductive learning from examples (Cohen et al., 1992), and several papers followed up this lead. The DLs considered were mostly sublanguages of CLASSIC which allowed for same-as equalities, i.e., expressions like (**same-as** (madeBy) (model $\circ$ madeBy)). Cohen et al. proposed an lcs algorithm for $\mathcal{ALN}$ and a language that allows for concept conjunction and same-as, which we will call $\mathcal{S}$. The algorithm for $\mathcal{S}$ was extended by Cohen and Hirsh (1994a) to CORE-CLASSIC, which additionally allows for value restrictions (see (Cohen & Hirsh, 1994b) for experimental results). Finally, Frazier and Pitt (1996) presented an lcs algorithm for full CLASSIC.

---

1. Observe that if the language already allows for disjunction, we have $lcs(C_1, \ldots, C_n) \equiv C_1 \sqcup \cdots \sqcup C_n$. In particular, this means that, for such languages, the lcs is not really of interest.

## 1.2 Total vs. Partial Attributes

In most knowledge representation systems, including DLs, functional relationships, here called *attributes* (also called "features" in the literature), are distinguished as a subclass of general relationships, at least in part because functional restrictions occur so frequently in practice[2]. In the above example, clearly madeBy and model are meant to be attributes, thus making unnecessary number restrictions like (**and** (**at-most** 1 madeBy) (**at-least** 1 madeBy)). In addition, distinguishing attributes helps identify tractable subsets of DL constructors: in CLASSIC, coreferences between attribute chains (as in the above examples) can be reasoned with efficiently (Borgida & Patel-Schneider, 1994), while if we changed to roles, e.g., allowed (**same-as** (repairs) (ownedBy ∘ repairsPaidFor)), the subsumption problem becomes undecidable (Schmidt-Schauß, 1989).

Whereas the distinction between roles and attributes in DLs is both theoretically and practically well understood, we have discovered that another distinction, namely the one between attributes being interpreted as total functions (*total attributes*) and those interpreted as partial functions (*partial attributes*), has "slipped through the cracks" of contemporary research. A total attribute always has a value in "the world out there", even if we do not know it in the knowledge base currently. A partial attribute may not have a value. This distinction is useful in practice, since there is a difference between a car possibly, but not necessarily, having a CD player, and the car necessarily having a manufacturer (which just may not be known in the current knowledge base). The latter is modeled by defining the attribute madeBy to be a total attribute. Note that with madeBy being a total attribute, *every* individual in the world of discourse (not only cars) must have a filler for madeBy. Since, however, no structural information is provided for fillers of madeBy of non-car individuals, all implications drawn about these fillers are trivial. Thus, making madeBy a total attribute seems reasonable in this case. A car's CD player, on the other hand, should be modeled by a partial attribute to express the fact that cars are not required to have a CD player. To indicate that a particular car does have a CD player, one would have to add the description (**at-least** 1 CDplayer).

## 1.3 New Results

As mentioned above, in conjunction with the same-as constructor, roles and attributes behave very differently with respect to subsumption. The main objective of this paper is to show that the distinction between total and partial attributes induces significantly different behaviour in computing the lcs, in the presence of **same-as**. More precisely, the purpose of this paper is twofold.

First, we show that with respect to the complexity of deciding subsumption there is no difference between partial and total attributes. Borgida and Patel-Schneider (1994) have shown that when attributes are total, subsumption of CLASSIC concept descriptions can be decided in polynomial time. As shown in the present work, slight modifications of the algorithm proposed by Borgida and Patel-Schneider suffice to handle partial attributes.

---

2. Readers coming from the Machine Learning community should be aware of the difference between our "attributes" (functional roles) and their "attributes", which are components of an input feature vector that usually describes an exemplar.

Moreover, these modifications do not change the complexity of the algorithm. Thus, partial and total attributes behave very similarly from the subsumption point of view.

Second, and this is the more surprising result of this paper, the distinction between partial and total attributes does have a significant impact on the problem of computing the lcs. Previous results on sublanguages of CLASSIC show that if partial attributes are used, the lcs of two concept descriptions always exists, and can be computed in polynomial time. If, however, only total attributes are involved, the situation is very different. The lcs need no longer even exist, and in case it exists its size may grow exponential in the size of the given concept descriptions. Nevertheless, the existence of the lcs of two concept descriptions can be decided in polynomial time.

Specifically, in previous work (Cohen et al., 1992; Cohen & Hirsh, 1994a; Frazier & Pitt, 1996) concerning the lcs computation in CLASSIC, constructions and proofs have been made without realizing the difference between the two types of attributes. Without going into details here, the main problem for lcs is that merely finite graphs have been employed, making the constructions applicable only for the partial attribute case. In addition to fixing these problems, this paper also presents the proper handling of inconsistent concepts in the lcs algorithm for CLASSIC presented by Frazier and Pitt (1996).

Although our results about subsumption are not as intriguing, the proofs to show the results on the lcs make extensive use of the corresponding subsumption algorithms, which is one reason we present them beforehand in this paper.

Returning to the general differences between the cases of total and partial attributes, one could say that the fundamental cause for the differences lies in the same-as constructor, whose semantics normally requires that (i) the two chains of attributes each have a value, and (ii) that these values coincide. In the case of total attributes, same-as obeys the principle

$$C \sqsubseteq u \downarrow v \text{ implies } C \sqsubseteq u \circ w \downarrow v \circ w$$

where $u$,$v$, and $w$ are sequences of total attributes, e.g., (madeBy ∘ model), because condition (i) is ensured by the total aspect of all the attributes. In the case of partial attributes, the above implication does not hold, because $w$, and hence $u \circ w$, is no longer guaranteed to have a value, implying that the same-as restriction may not hold. Clearly, this implication affects the results of subsumption. As far as lcs is concerned, a certain graph (representing the lcs of the two given concepts) may be infinite in the case of total attributes, thus jeopardizing the existence of the lcs.

The more general significance of our result is that knowledge representation language designers *and* users need to explicitly check at the beginning whether they deal with total or partial attributes because the choice can have significant effects. Although in some situations total attributes are convenient, to guarantee the existence of attributes without having to resort to number restrictions, our results show that they can have drawbacks. All things considered, requiring all attributes to be total appears to be less desirable. Concerning CLASSIC, the technical results in this paper support the use of partial attributes because these ensure the existence of the lcs and its computation in polynomial time as well as the efficient decision of subsumption. Moreover, the current implementation of the CLASSIC subsumption algorithm does not require major changes in order to handle partial attributes.

The outline of this paper is as follows: In the following section, the basic notions necessary for our investigations are introduced. Then, in the two subsequent sections, subsumption and lcs computation in CLASSIC with partial attributes is investigated. More precisely, in Section 3 we offer a subsumption algorithm for the sublanguage CLASSIC⁻ of CLASSIC, which contains all main CLASSIC-constructors; in Section 4, we present an lcs algorithm for CLASSIC⁻ concept descriptions, along the lines of that proposed by Cohen and Hirsh (1994a), and formally prove its correctness, thereby resolving some shortcomings of previous lcs algorithms, which did not handle inconsistencies properly. Finally, Section 5 covers the central new result of this paper, i.e., the lcs computation in presence of total attributes. For this section, we restrict our investigations to the sublanguage $\mathcal{S}$ of CLASSIC⁻ in order to concentrate on the changes caused by going from partial to total attributes. Nevertheless, we strongly conjecture that all the results proved in this section can easily be extended to CLASSIC⁻ and CLASSIC using similar techniques as the one employed in the two previous sections.

## 2. Formal Preliminaries

In this section, we introduce the syntax and semantics of the description languages considered in this paper and formally define subsumption and equivalence of concept descriptions. Finally, the least common subsumer of concept descriptions is specified.

**Definition 1** *Let $\mathcal{C}$, $\mathcal{R}$, and $\mathcal{A}$ be disjoint finite sets representing the set of concept names, the set of role names, and the set of attribute names. The set of all CLASSIC⁻-concept descriptions over $\mathcal{C}$, $\mathcal{R}$, and $\mathcal{A}$ is inductively defined as follows:*

- *Every element of $\mathcal{C}$ is a concept description* (concept name, like Car).

- *The symbol $\top$ is a concept description* (top concept, denoting the universe of all objects).

- *If $r \in \mathcal{R}$ is a role and $n \geq 0$ is a nonnegative integer, then $\leq n\, r$ and $\geq n\, r$ are concept descriptions* (number restrictions, like $\geq 10$ repairs).

- *If $C$ and $D$ are concept descriptions, then $C \sqcap D$ is a concept description* (concept conjunction).

- *If $C$ is a concept description and $r$ is a role or an attribute, then $\forall r.C$ is a concept description* (value restriction, like ∀madeBy.Manufacturer).

- *If $k, h \geq 0$ are non-negative integers and $a_1, \ldots, a_k, b_1, \ldots, b_h \in \mathcal{A}$ are attributes, then $a_1 \circ \cdots \circ a_k \downarrow b_1 \circ \cdots \circ b_h$ is a concept description* (same-as equality, like madeBy ↓ model ∘ madeBy). *Note that the two sequences may be empty, i.e., $k = 0$ or $h = 0$. The empty sequence is denoted by $\varepsilon$.*

Often we dispense with $\circ$ in the composition of attributes. For example, the sequence $a_1 \circ \cdots \circ a_k$ is simply written as $a_1 \cdots a_k$. Moreover, we will use $\forall r_1 \cdots r_n.C$ as abbreviation of $\forall r_1.\forall r_2 \cdots \forall r_n.C$, where we have $\forall \varepsilon.C$ in case $n = 0$, and this denotes $C$.

As usual, the semantics of CLASSIC⁻ is defined in a model-theoretic way by means of interpretations.

**Definition 2** *An* interpretation $\mathcal{I}$ *consists of a nonempty domain* $\Delta^{\mathcal{I}}$ *and an interpretation function* $\cdot^{\mathcal{I}}$. *The interpretation function assigns extensions to atomic identifiers as follows:*

- *The extension of a concept name* $E$ *is some subset* $E^{\mathcal{I}}$ *of the domain.*

- *The extension of a role name* $r$ *is some subset* $r^{\mathcal{I}}$ *of* $\Delta^{\mathcal{I}} \times \Delta^{\mathcal{I}}$.

- *The extension of an attribute name* $a$ *is some partial function* $a^{\mathcal{I}}$ *from* $\Delta^{\mathcal{I}}$ *to* $\Delta^{\mathcal{I}}$, *i.e., if* $(x, y_1) \in a^{\mathcal{I}}$ *and* $(x, y_2) \in a^{\mathcal{I}}$ *then* $y_1 = y_2$.

*Given roles or attributes* $r_i$, *we use* $(r_1 \cdots r_n)^{\mathcal{I}}$ *to denote the composition of the binary relations* $r_i^{\mathcal{I}}$. *If* $n = 0$ *then the result is* $\varepsilon^{\mathcal{I}}$, *which denotes the identity relation, i.e.,* $\varepsilon^{\mathcal{I}} := \{(d, d) \mid d \in \Delta^{\mathcal{I}}\}$. *For an individual* $d \in \Delta^{\mathcal{I}}$, *we define* $r^{\mathcal{I}}(d) := \{e \mid (d, e) \in r^{\mathcal{I}}\}$. *If the* $r_i$'s *are attributes, we say that* $(r_1 \cdots r_n)^{\mathcal{I}}$ *is* defined *for* $d$ *iff* $(r_1 \cdots r_n)^{\mathcal{I}}(d) \neq \emptyset$; *occasionally, we will refer to* $(r_1 \cdots r_n)(d)^{\mathcal{I}}$ *as the image of* $d$ *under* $(r_1 \cdots r_n)^{\mathcal{I}}(d)$.

*The extension* $C^{\mathcal{I}}$ *of a concept description* $C$ *is inductively defined as follows:*

- $\top^{\mathcal{I}} := \Delta^{\mathcal{I}}$;

- $(\geq n\, r)^{\mathcal{I}} := \{d \in \Delta^{\mathcal{I}} \mid cardinality(\{e \in \Delta^{\mathcal{I}} \mid (d, e) \in r^{\mathcal{I}}\}) \geq n\}$;

- $(\leq n\, r)^{\mathcal{I}} := \{d \in \Delta^{\mathcal{I}} \mid cardinality(\{e \in \Delta^{\mathcal{I}} \mid (d, e) \in r^{\mathcal{I}}\}) \leq n\}$;

- $(C \sqcap D)^{\mathcal{I}} := C^{\mathcal{I}} \cap D^{\mathcal{I}}$;

- $(\forall r.C)^{\mathcal{I}} := \{d \in \Delta^{\mathcal{I}} \mid r^{\mathcal{I}}(d) \subseteq C^{\mathcal{I}}\}$ *where* $r$ *is a role or an attribute;*

- $(a_1 \cdots a_k \downarrow b_1 \cdots b_h)^{\mathcal{I}} := \{d \in \Delta^{\mathcal{I}} \mid (a_1 \cdots a_k)^{\mathcal{I}} \text{ and } (b_1 \cdots b_h)^{\mathcal{I}} \text{ are defined for } d$
  $\text{and } (a_1 \cdots a_k)^{\mathcal{I}}(d) = (b_1 \cdots b_h)^{\mathcal{I}}(d)\}$.

Note that in the above definition attributes are interpreted as partial functions. Since the main point of this paper is to demonstrate the impact of different semantics for attributes, we occasionally restrict the set of interpretations to those that map attributes to *total* functions. Such interpretations are called *t-interpretations* and the attributes interpreted in this way are called *total attributes* in order to distinguish them from *partial* ones.

We stress, as remarked in the introduction, that in the definition of $(a_1 \cdots a_k \downarrow b_1 \cdots b_h)^{\mathcal{I}}$, $a_1 \cdots a_k$ and $b_1 \cdots b_h$ must be defined on $d$ in order for $d$ to satisfy the same-as restriction. Although this is the standard semantics for same-as equalities, one could also think of relaxing this restriction. For example, the same-as condition might be specified to hold if *either* both paths are undefined *or* both images are defined and have identical values. A third definition might be satisfied if even just one of the paths is undefined. Each of these definitions of the semantics of same-as might lead to different results. However, in this paper we only pursue the standard semantics.

The subsumption relationship between concept descriptions is defined as follows.

**Definition 3** *A concept description* $C$ *is subsumed by the concept description* $D$ ($C \sqsubseteq D$ *for short) if and only if for all interpretations* $\mathcal{I}$, $C^{\mathcal{I}} \subseteq D^{\mathcal{I}}$. *If we consider only total interpretations, we get* t-subsumption: $C \sqsubseteq_t D$ *iff* $C^{\mathcal{I}} \subseteq D^{\mathcal{I}}$ *for all* $t$-interpretations $\mathcal{I}$.

Having defined subsumption, equivalence of concept descriptions is defined in the usual way: $C \equiv D$ if and only if $C \sqsubseteq D$ and $D \sqsubseteq C$. T-equivalence $C \equiv_t D$ is specified analogously.

As already mentioned in the introduction, the main difference between partial and total attributes with respect to subsumption is that $u \downarrow v \sqsubseteq_t u \circ w \downarrow v \circ w$ holds for all attribute chains $u, v, w$, whereas it is not necessarily the case that $u \downarrow v \sqsubseteq u \circ w \downarrow v \circ w$.

Finally, before introducing the lcs operation formally and concluding this section, we comment on the expressive power of CLASSIC⁻, since (syntactically) CLASSIC⁻ lacks some common constructors. Although CLASSIC⁻, as introduced here, does not contain the *bottom concept* $\bot$ explicitly, it can be expressed by, e.g., $(\geq 1\, r) \sqcap (\leq 0\, r)$. We will use $\bot$ as an abbreviation for inconsistent concept descriptions. Furthermore, *primitive negation*, i.e., negation of concept names, can be simulated by number restrictions. For a concept name $E$ one can replace every occurrence of $E$ by $(\geq 1\, r_E)$ and the negation $\neg E$ of $E$ by $(\leq 0\, r_E)$ where $r_E$ is a new role name. Finally, for an attribute $a$ the following equivalences hold: $(\geq n\, a) \equiv \bot$ for $n \geq 2$; $(\geq 1\, a) \equiv (a \downarrow a)$; $(\geq 0\, a) \equiv \top$; $(\leq n\, a) \equiv \top$ for $n \geq 1$; and $(\leq 0\, a) \equiv (\forall a.\bot)$. These show that we do not lose any expressive power by not allowing for number restrictions on attributes. Still, full CLASSIC is somewhat more expressive than CLASSIC⁻. This is mainly due to the introduction of individuals (also called nominals) in CLASSIC. For the sake of completeness we give the syntax of the full CLASSIC language.[3] This requires a further set, $\mathcal{O}$, representing the set of individual names. Then we can define two additional concept constructors

- $\{e_1, ..., e_m\}$, for individuals $e_i \in \mathcal{O}$ (*enumeration* as in $\{Fall, Summer, Spring\}$)

- $p : e$ for a role or attribute $p$, and an individual $e$ (*fills* as in $currentSeason : Summer$).

In a technical report, Küsters and Borgida (1999) extend some of the results presented in this work to full CLASSIC, in the case when individuals have a non-standard semantics.

The least common subsumer of a set of concept descriptions is the most specific concept subsuming all concept descriptions of the set:

**Definition 4** *The concept description $D$ is the* least common subsumer (lcs) *of the concept descriptions $C_1, \ldots, C_n$ ($lcs(C_1, \ldots, C_n)$ for short) iff i) $C_i \sqsubseteq D$ for all $i = 1, \ldots, n$ and ii) for every $D'$ with that property $D \sqsubseteq D'$. Analogously, we define $lcs_t(C_1, \ldots, C_n)$ using $\sqsubseteq_t$ instead of $\sqsubseteq$.*

Note that the lcs of concept descriptions may not exist, but if it does, by definition it is uniquely determined up to equivalence. In this sense, we may refer to *the* lcs.

In the following two sections, attributes are always interpreted as partial functions; only in Section 5 do we consider total attributes.

## 3. Characterizing Subsumption in CLASSIC⁻

In this section we modify the characterization of t-subsumption for CLASSIC, as proposed by Borgida and Patel-Schneider (1994), to handle the case of partial attributes. We do so in detail, because the tools used for deciding subsumption are intimately related to the computation of lcs.

---

3. Even here we are omitting constructs dealing with integers and other so-called "host individuals", which cannot have roles of their own and can only act as role/attribute fillers.

T-subsumption in CLASSIC is decided by a multi-part process. First, descriptions are turned into description graphs. Next, description graphs are put into canonical form, where certain inferences are explicated and other redundancies are reduced by combining nodes and edges in the graph. Finally, t-subsumption is determined between a description and a canonical description graph.

In order to "inherit" the proofs, we have tried to minimize the necessary adjustments to the specification in (Borgida & Patel-Schneider, 1994). For this reason, roughly speaking, attributes are treated as roles unless they form part of a same-as equality. (Note that attributes participating in a same-as construct must have values!) To some extent, this will allow us to adopt the semantics of the original description graphs, which is crucial for proofs. However, the two different occurrences of attributes, namely, in a same-as equality vs. a role in a value-restriction, require us to modify and extend the definition of description graphs, the normalization rules, and the subsumption algorithm itself.

In the following, we present the steps of the subsumption algorithm in detail. We start with the definition of description graphs.

## 3.1 Description Graphs

Intuitively, description graphs reflect the syntactic structure of concept descriptions. A description graph is a labeled, directed multigraph, with a distinguished node. Roughly speaking, the edges (*a-edges*) of the graph capture the constraints expressed by same-as equalities. The labels of nodes contain, among others, a set of so-called *r-edges*, which correspond to value restrictions. Unlike the description graphs defined by Borgida and Patel-Schneider, here the r-edges are not only labeled with role names but also with attribute names. (We shall comment later on the advantage of this modification in order to deal with partial attributes.) The r-edges lead to *nested description graphs*, representing the concepts of the corresponding value restrictions.

Before defining description graphs formally, in Figure 1 we present a graph corresponding to the concept description Lemon defined in the introduction. We use $G(\mathsf{Manufacturer})$, $G(\mathsf{Model})$, as well as $G(\mathsf{RepairReport})$ to denote description graphs for the concept names Manufacturer, Model, and RepairReport. These graphs are very simple; they merely consist of one node, labeled with the corresponding concept name. In general, such graphs can be more complex since a value restriction like $\forall r.C$ leads to a (possibly complex) nested concept description $C$.

Although number restrictions on attributes are not allowed, r-edges labeled with attributes, like model and madeBy, always have the restriction $[0, 1]$ in order to capture the semantics of attributes. Formally, description graphs, nodes, and edges are defined mutually recursively as follows:

**Definition 5** *A description graph $G$ is a tuple $(N, E, n_0, l)$, consisting of a finite set $N$ of nodes; a finite set $E$ of edges (a-edges); a distinguished node $n_0 \in N$ (root of the graph); and a function $l$ from $N$ into the set of labels of nodes. We will occasionally use the notation $G.Nodes$, $G.Edges$, and $G.root$ to access the components $N$, $E$ and $n_0$ of the graph $G$.*

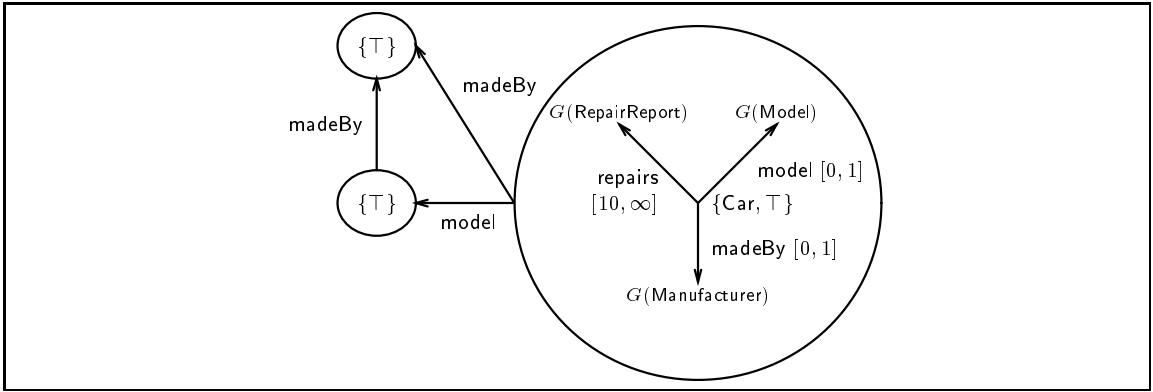

Figure 1: A description graph for Lemon, where the large node is the root of the graph

*An* a-edge *is a tuple of the form* $(n_1, a, n_2)$ *where* $n_1$, $n_2$ *are nodes and* $a$ *is an attribute name.*

*A label of a node is defined to be* $\bot$ *or a tuple of the form* $(C, H)$, *consisting of a finite set* $C$ *of concept names (the* atoms *of the node) and a finite set* $H$ *of tuples (the* r-edges *of the node). Concept names in a description graph stand for atomic concept names and* $\top$. *We will occasionally use the notation* $n.Atoms$ *and* $n.REdges$ *to access the components* $C$ *and* $H$ *of the node* $n$.

*An* r-edge *is a tuple,* $(r, m, M, G')$, *consisting of a role or attribute name,* $r$; *a min,* $m$, *which is a non-negative integer; a max,* $M$, *which is a non-negative integer or* $\infty$; *and a (recursively nested) description graph* $G'$. *The graph* $G'$ *will often be called the* restriction graph *of the node for the role* $r$. *We require the nodes of* $G'$ *to be distinct from all the nodes of* $G$ *and other nested description graphs of* $G$. *If* $r$ *is an attribute, then we require:* $m = 0$ *and* $M \in \{0, 1\}$.

Given a description graph $G$ and a node $n \in G.Nodes$, we define $G_{|n}$ to be the graph $(N, E, n, l)$; $G_{|n}$ is said to be rooted at $n$. A sequence $p = n_0 a_1 a_2 \cdots a_k n_k$ with $k \geq 0$ and $(n_{i-1}, a_i, n_i) \in G.Edges$, $i = 1, \ldots, k$, is called *path in* $G$ *from the node* $n_0$ *to* $n_k$ ($p \in G$ for short); for $k = 0$ the path $p$ is called *empty*; $w = a_1 \cdots a_k$ is called the *label* of $p$ (the empty path has label $\varepsilon$); $p$ is called *rooted* if $n_0$ is the root of $G$. Occasionally, we write $n_0 a_1 \cdots a_k n_k \in G$ omitting the intermediate nodes.

Throughout this work we make the assumption that *description graphs are connected*. A description graph is said to be *connected* if all nodes of the graph can be reached by a rooted path and all nested graphs are connected. The semantics of description graphs (see Definition 6) is not altered if nodes that cannot be reached from the root are deleted.

In order to merge description graphs we need the notion of "recursive set of nodes" of a description graph $G$: The *recursive set of nodes of* $G$ is the union of the nodes of $G$ and the recursive set of nodes of all nested description graphs of $G$.

Just as for concept descriptions, the semantics of description graphs is defined by means of an interpretation $\mathcal{I}$. We introduce a function $\Upsilon$ which assigns an individual of the domain of $\mathcal{I}$ to every node of the graph. This ensures that all same-as equalities are satisfied.

**Definition 6** *Let $G = (N, E, n_0, l)$ be a description graph and let $\mathcal{I}$ be an interpretation.*

*An element, $d$, of $\Delta^{\mathcal{I}}$ is in $G^{\mathcal{I}}$, iff there is some total function, $\Upsilon$, from $N$ into $\Delta^{\mathcal{I}}$ such that*

1. $d = \Upsilon(n_0)$;

2. *for all $n \in N$, $\Upsilon(n) \in n^{\mathcal{I}}$; and*

3. *for all $(n_1, a, n_2) \in E$ we have $(\Upsilon(n_1), \Upsilon(n_2)) \in a^{\mathcal{I}}$.*

*The extension $n^{\mathcal{I}}$ of a node $n$ with label $\perp$ is the empty set. An element, $d$, of $\Delta^{\mathcal{I}}$ is in $n^{\mathcal{I}}$, where $l(n) = (C, H)$, iff*

1. *for all $B \in C$, we have $d \in B^{\mathcal{I}}$; and*

2. *for all $(r, m, M, G') \in H$,*

   (a) *there are between $m$ and $M$ elements, $d'$, of the domain such that $(d, d') \in r^{\mathcal{I}}$; and*

   (b) *$d' \in G'^{\mathcal{I}}$ for all $d'$ such that $(d, d') \in r^{\mathcal{I}}$.*

Cohen and Hirsh (1994a) defined the semantics of description graphs in a different way, avoiding the introduction of a total function $\Upsilon$. The problem with their definition is, however, that it is only well-defined for acyclic graphs, which, for example, excludes same-as equalities of the form $\varepsilon \downarrow spouse \circ spouse$, or even $p \downarrow p \circ q$.

The semantics of the graphs proposed by Borgida and Patel-Schneider (1994) is similar to Definition 6. However, in that paper a-edges captured not only same-as equalities but also *all* value restrictions on attributes. Still, in the context of partial attributes, we could not define the semantics of description graphs by means of a total function $\Upsilon$ since some attributes might not have fillers. Specifying the semantics of description graphs in terms of *partial* mappings $\Upsilon$ would make the definition even longer. Furthermore, the proofs in (Borgida & Patel-Schneider, 1994) would not carry over as easily. Therefore, in order to keep $\Upsilon$ a total function, value restrictions of attributes are initially always translated into r-edges. The next section will present the translation of concept descriptions into description graphs in detail.

Having defined the semantics of description graphs, subsumption and equivalence between description graphs (e.g., $H \sqsubseteq G$) as well as concept descriptions and description graphs (e.g., $C \sqsubseteq G$) is defined in the same way as subsumption and equivalence between concept descriptions.

## 3.2 Translating Concept Descriptions into Description Graphs

Following Borgida and Patel-Schneider (1994), a CLASSIC$^-$ concept description is turned into a description graph by a recursive process. In this process, nodes and description graphs are often merged.

**Definition 7** *The merge of two nodes, $n_1 \oplus n_2$, is a new node $n$ with the following label: if $n_1$ or $n_2$ have label $\perp$, then the label of $n$ is $\perp$. Otherwise if both labels are not equal to $\perp$, then $n.Atoms = n_1.Atoms \cup n_2.Atoms$ and $n.REdges = n_1.REdges \cup n_2.REdges$.*

If $G_1 = (N_1, E_1, n_1, l_1)$ and $G_2 = (N_2, E_2, n_2, l_2)$ are two description graphs with disjoint recursive sets of nodes, then the merge of $G_1$ and $G_2$, $G := G_1 \oplus G_2 = (N, E, n_0, l)$, is defined as follows:

1. $n_0 := n_1 \oplus n_2$;

2. $N := (N_1 \cup N_2 \cup \{n_0\}) \setminus \{n_1, n_2\}$;

3. $E := (E_1 \cup E_2)[n_1/n_0, n_2/n_0]$, i.e., $E$ is the union of $E_1$ and $E_2$ where every occurrence of $n_1, n_2$ is substituted by $n_0$;

4. $l(n) := l_1(n)$ for all $n \in N_1 \setminus \{n_1\}$; $l(n) := l_2(n)$ for all $n \in N_2 \setminus \{n_2\}$; and $l(n_0)$ is defined by the label obtained by merging $n_1$ and $n_2$.

Now, a CLASSIC$^-$-concept description $C$ can be turned into its *corresponding description graph* $G(C)$ by the following translation rules.

1. $\top$ is turned into a description graph with one node $n_0$ and no a-edges. The only atom of the node is $\top$ and the set of r-edges is empty.

2. A concept name is turned into a description graph with one node and no a-edges. The atoms of the node contain only the concept name and the node has no r-edges.

3. A description of the form $(\geq n\, r)$ is turned into a description graph with one node and no a-edges. The node has as its atoms $\top$ and it has a single r-edge $(r, n, \infty, G(\top))$ where $G(\top)$ is specified by the first translation rule.

4. A description of the form $(\leq n\, r)$ is turned into a description graph with one node and no a-edges. The node has as its atom $\top$ and it has a single r-edge $(r, 0, n, G(\top))$.

5. A description of the form $a_1 \cdots a_p \downarrow b_1 \cdots b_q$ is turned into a graph with pairwise distinct nodes $n_1, \ldots, n_{p-1}, m_1, \ldots, m_{q-1}$, the root $m_0 := n_0$, and an additional node $n_p = m_q := n$; the set of a-edges consists of $(n_0, a_1, n_1)$, $(n_1, a_2, n_2), \ldots, (n_{p-1}, a_p, n_p)$ and $(m_0, b_1, m_1)$, $(m_1, b_2, m_2), \ldots, (m_{q-1}, b_q, m_q)$, i.e., two disjoint paths which coincide on their starting point, $n_0$, and their final point, $n$. (Note that for $p = 0$ the first path is the empty path from $n_0$ to $n_0$ and for $q = 0$ the second path is the empty path from $n_0$ to $n_0$.) All nodes have $\top$ as their only atom and no r-edges.

6. A description of the form $\forall r.C$, where $r$ is a role, is turned into a description graph with one node and no a-edges. The node has the atom $\{\top\}$ and it has a single r-edge $(r, 0, \infty, G(C))$.

7. A description of the form $\forall a.C$, where $a$ is an attribute, is turned into a description graph with one node and no a-edges. The node has the atom $\{\top\}$ and it has a single r-edge $(a, 0, 1, G(C))$. (In the work by Borgida and Patel-Schneider, the concept description $\forall a.C$ is turned into an a-edge. As already mentioned, this would cause problems for attributes interpreted as partial functions when defining the semantics by means of $\Upsilon$ as specified in Definition 6.)

8. To turn a description of the form $C \sqcap D$ into a description graph, construct $G(C)$ and $G(D)$ and merge them.

Figure 1 shows the description graph built in this way for the concept Lemon of our example. It can easily be verified that the translation preserves extensions:

**Theorem 1** *A concept description $C$ and its corresponding description graph $G(C)$ are equivalent, i.e.,$C^{\mathcal{I}} = G(C)^{\mathcal{I}}$ for every interpretation $\mathcal{I}$.*

The main difficulty in the proof of this theorem is in showing that merging two description graphs corresponds to the conjunction of concept descriptions.

**Lemma 1** *For all interpretations $\mathcal{I}$, if $n_1$ and $n_2$ are nodes, then $(n_1 \oplus n_2)^{\mathcal{I}} = n_1^{\mathcal{I}} \cap n_2^{\mathcal{I}}$; if $G_1$ and $G_2$ are description graphs then $(G_1 \oplus G_2)^{\mathcal{I}} = G_1^{\mathcal{I}} \cap G_2^{\mathcal{I}}$.*

The proof of the preceding statement is rather simple and like the one in (Borgida & Patel-Schneider, 1994).

### 3.3 Translating Description Graphs to Concept Descriptions

Although the characterization of subsumption does not require translating description graphs to concept descriptions, this translation is presented here to show that concept descriptions and description graphs are equivalent representations of CLASSIC⁻ concept descriptions. In subsequent sections, we will in fact need to turn graphs into concept descriptions.

The translation of a description graph $G$ can be specified in a rather straightforward recursive definition. The main idea of the translation stems from Cohen and Hirsh (1994a), who employed spanning trees to translate same-as equalities. A *spanning tree* of a (connected) graph is a tree rooted at the same node as the graph and containing all nodes of the graph. In particular, it coincides with the graph except that some a-edges are deleted. For example, one possible spanning tree $T$ for $G$ in Figure 1 is obtained by deleting the a-edge labeled madeBy, whose origin is the root of $G$.

Now, let $G$ be a connected description graph and $T$ be a spanning tree for it. Then, the corresponding concept description $C_G$ is obtained as a conjunction of the following descriptions:

1. $C_G$ contains (i) a same-as equality $v \downarrow v$ for every leaf $n$ of $T$, where $v$ is the label of the rooted path in $T$ to $n$; and (ii) a same-as equality $v_1 \circ a \downarrow v_2$ for each a-edge $(n_1, a, n_2) \in G.Edges$ not contained in $T$, where $v_i$ is the label of the rooted path to $n_i$ in $T$, $i = 1, 2$.

2. for every node $n$ in $T$, $C_G$ contains a value restriction $\forall v.C_n$, where $v$ is the label of the rooted path in $T$ to $n$, and $C_n$ denotes the translation of the label of $n$, i.e., $C_n$ is a conjunction obtained as follows:

   - every concept name in the atoms of $n$ is a conjunct in $C_n$;
   - for every r-edge $(r, m, M, G')$ of $n$, $C_n$ contains (a) the number restrictions $(\geq m r)$ and $(\leq M r)$ (in case $r$ is a role and $M \neq \infty$) and (b) the value restriction $\forall r.C_{G'}$, where $C_{G'}$ is the recursively defined translation of $G'$.

179

In case the set of atoms and r-edges of $n$ is empty, define $C_n := \top$.

Referring to the graph $G$ in Figure 1, $C_G$ contains the same-as equalities model ∘ madeBy ↓ model ∘ madeBy and madeBy ↓ model ∘ madeBy. Furthermore, if $n_0$ denotes the root of $G$, $C_G$ has the value restrictions $\forall \varepsilon.C_{n_0}$, $\forall$model.$\top$, and $\forall$model madeBy.$\top$, where $C_{n_0}$ corresponds to Lemon as defined in the introduction, but without the same-as equality. Note that, although in this case the same-as equality model ∘ madeBy ↓ model ∘ madeBy is not needed, one cannot dispense with 1.(i) in the construction above, as illustrated by the following example: Without 1.(i), the description graph $G(a \downarrow a)$ would be turned into the description $\top$, which is not equivalent to $a \downarrow a$ since the same-as equality requires that the path $a$ has a value, which may not be the case.

It is easy to prove that the translation thus defined is correct in the following sense (Küsters & Borgida, 1999).

**Lemma 2** *Every connected description graph $G$ is equivalent to its translation $C_G$, i.e., for all interpretations $\mathcal{I}$: $G^{\mathcal{I}} = C_G^{\mathcal{I}}$.*

## 3.4 Canonical Description Graphs

In the following we occasionally refer to "*marking a node incoherent*"; this means that the label of this node is changed to $\bot$. "*Marking a description graph as incoherent*" means that the description graph is replaced by the graph $G(\bot)$ corresponding to $\bot$, i.e., the graph consisting only of one node with label $\bot$.

One important property of canonical description graphs is that they are deterministic, i.e., every node has at most one outgoing edge (a-edge or r-edge) labeled with the same attribute or role name. Following Borgida and Patel-Schneider (1994), in order to turn a description graph into a canonical graph we need to merge a-edges and r-edges. In addition, different from their work, it might be necessary to "lift" r-edges to a-edges.

To *merge two a-edges* $(n, a, n_1)$ and $(n, a, n_2)$ in a description graph $G$, replace them with a single new edge $(n, a, n')$ where $n'$ is the result of merging $n_1$ and $n_2$. In addition, replace $n_1$ and $n_2$ by $n'$ in all other a-edges of $G$.

In order to *merge two r-edges* $(r, s_1, k_1, G_1)$, $(r, s_2, k_2, G_2)$ replace them by the new r-edge $(r, max(s_1, s_2), min(k_1, k_2), G_1 \oplus G_2)$.

To *lift up* an r-edge $(a, m, M, G_a)$ of a node $n$ in a concept graph $G$ with an a-edge $(n, a, n_1)$, remove it from $n.REdges$, and augment $G$ by adding $G_a.Nodes$ to $G.Nodes$, $G_a.Edges$ to $G.Edges$, as well as adding $(n, a, G_a.Root)$ to $G.Edges$. A *precondition* for applying this transformation is that $M = 1$, or $M = 0$ and $G_a$ corresponds to the graph $G(\bot)$. The reason for this precondition is that if an r-edge of the form $(a, 0, 0, G_a)$ is lifted without $G_a$ being inconsistent, the fact that no $a$-successors are allowed is lost. Normalization rule 5 (see below) will guarantee that this precondition can always be satisfied.

A description graph $G$ is transformed into *canonical form* by exhaustively applying the following *normalization rules*. A graph is called *canonical* if none of these rules can be applied.

1. If some node in $G$ is marked incoherent, mark the description graph as incoherent. (*Reason: Even if the node is not a root, attributes corresponding to a-edges must always* *have a value (since they participate in same-as equalities), and this value cannot belong to the empty set.)*

2. If some r-edge in a node has its min greater than its max, mark the node incoherent. *(Reason: $\geq 2\, r \sqcap \leq 1\, r \equiv \bot$)*

3. Add $\top$ to the atoms of every node, if absent.

4. If some r-edge in a node has its restriction graph marked incoherent, change its max to 0. *(Reason: $(\leq 0\, r) \equiv \forall r.\bot$.)*

5. If some r-edge in a node has a max of 0, mark its restriction graph as incoherent. *(Reason: See 4.)*

6. If some r-edge is of the form $(r, 0, \infty, G')$ where $G'$ only contains one node with empty set of atoms or with the atoms set to $\{\top\}$ and no r-edges, then remove this r-edge. *(Reason: $\forall r.\top \equiv \top$.)*

7. If some node has two r-edges labeled with the same role, merge the two edges, as described above. *(Reason: $\forall r.C \sqcap \forall r.D \equiv \forall r.(C \sqcap D)$.)*

8. If some description graph has two a-edges from the same node labeled with the same attribute, merge the two edges, as described above. *(Reason: $\forall a.C \sqcap \forall a.D \equiv \forall a.(C \sqcap D)$.)*

9. If some node in a graph has both an a-edge and an r-edge for the same attribute, then "lift up the r-edge" if the precondition is satisfied (see above). *(Reason: The value restrictions imposed on attributes that participate in same-as equalities must be made explicit and gathered at one place similar to the previous to cases.)*

We need to show that the transformations to canonical form do not change the semantics of the graph. The main difficulty is in showing that the merging processes and the lifting preserve the semantics. The only difference from (Borgida & Patel-Schneider, 1994) is that in addition to merging r-edges and a-edges we also need to lift up r-edges. Therefore, we omit the proofs showing that merging edges preserves extensions. The proofs of the following two lemmas are routine and quite similar to the one of Lemma 5.

**Lemma 3** *Let $G = (N, E, n_0, l)$ be a description graph with two mergeable a-edges and let $G' = (N', E', n', l')$ be the result of merging these two a-edges. Then, $G \equiv G'$.*

**Lemma 4** *Let $n$ be a node with two mergeable r-edges and let $n'$ be the node with these edges merged. Then, $n^{\mathcal{I}} = n'^{\mathcal{I}}$ for every interpretation $\mathcal{I}$.*

**Lemma 5** *Let $G = (N, E, n_0, l)$ be a description graph with node $n$ and a-edge $(n, a, n'')$. Suppose $n$ has an associated r-edge $(a, m, M, G_a)$. Provided that the precondition for lifting r-edges is satisfied and that $G' = (N', E', n', l')$ is the result of this transformation, then $G \equiv G'$.*
**Proof.** It is sufficient to show that $G^{\mathcal{I}}_{|n} = G'^{\mathcal{I}}_{|n}$, since only the label of $n$ is changed in $G'$ and only $n$ obtains an additional a-edge, which points to the graph $G_a$ not connected to

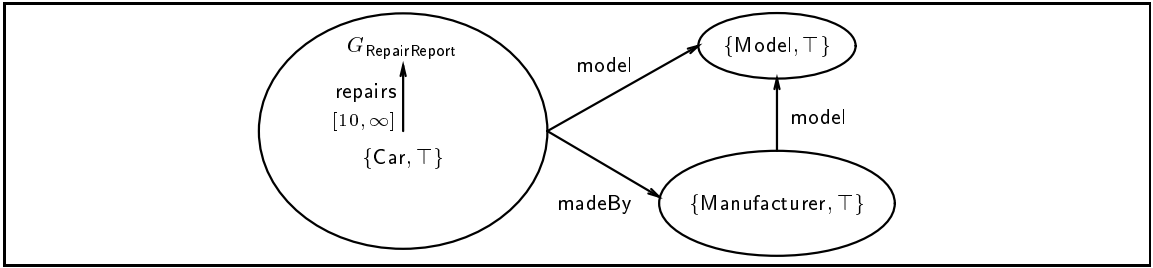

Figure 2: The canonical description graph for Lemon, where the left-most node is the root.

the rest of $G'$. W.l.o.g. we therefore may assume that $n$ is the root of $G$, i.e., $n = n_0$. Let $d \in G^{\mathcal{I}}$. Thus, there is a function $\Upsilon$ from $N$ into $\Delta^{\mathcal{I}}$ as specified in Definition 6 and an individual $e$ such that $d = \Upsilon(n)$, $e = \Upsilon(n'')$, and $(d, e) \in a^{\mathcal{I}}$. This implies $e \in G_a^{\mathcal{I}}$. Hence, there exists a function $\Upsilon'$ from $G_a.Nodes$ into $\Delta^{\mathcal{I}}$ for $G_a$ and $e$ satisfying the conditions in Definition 6. Since the sets of nodes of $G$ and $G_a$ are disjoint, we can define $\Upsilon''$ to be the union of $\Upsilon$ and $\Upsilon'$, i.e., $\Upsilon''(m) := \Upsilon(m)$ for all nodes $m$ in $G$ and $\Upsilon''(m) := \Upsilon'(m)$ for all nodes $m$ in $G_a$. Since, by construction, for the additional a-edge $(n, a, G_a.Root) \in E'$ we have $(\Upsilon''(n), \Upsilon''(G_a.Root)) \in a^{\mathcal{I}}$, it follows that all conditions in Definition 6 are satisfied for $d$ and $G'$, and thus, $d \in G'^{\mathcal{I}}$.

Now let $d \in G'^{\mathcal{I}}$. Thus, there is a function $\Upsilon''$ from $N'$ into $\Delta^{\mathcal{I}}$ according to Definition 6. Let $e := \Upsilon''(G_a.Root) = \Upsilon''(n'')$. Let $G''$ be the description graph we obtain from $G'$ by deleting the nodes corresponding to $G_a$, which is the same graph as $G$ without the r-edge $(a, m, M, G_a)$. If we restrict $\Upsilon''$ to the nodes of $G''$, then it follows $d \in G''^{\mathcal{I}}$. Furthermore, restricting $\Upsilon''$ to the nodes of $G_a$ yields $e \in G_a^{\mathcal{I}}$. In particular, $G_a$ can not be marked incoherent. Then, our precondition ensures $M = 1$. Thus, since $e$ is the only $a$-successor of $d$, we can conclude $d \in G^{\mathcal{I}}$. □

Having dealt with the issue of merging and lifting, it is now easy to verify that "normalization" does not affect the meaning of description graphs.

**Theorem 2** *If $G$ is a description graph and $G'$ is the corresponding canonical description graph, then $G \equiv G'$.*

As an example, the canonical description graph of the graph given in Figure 1 is depicted in Figure 2.

## 3.5 Subsumption Algorithm

The final part of the subsumption process is checking to see if a canonical description graph is subsumed by a concept description. As in Borgida and Patel-Schneider (1994), where attributes are total, it turns out that it is not necessary to turn the potential subsumer into a canonical description graph. The subsumption algorithm presented next can also be considered as a characterization of subsumption.

**Algorithm 1 (Subsumption Algorithm)** *Given a concept description $D$ and description graph $G = (N, E, n_0, l)$, subsumes?$(D, G)$ is defined to be true if and only if one of the following conditions hold:*

1. *The description graph $G$ is marked incoherent.*

2. *$D$ is a concept name or $\top$, and $D$ is an element of the atoms of $n_0$.*

3. *$D$ is $(\geq n\, r)$ and i) some $r$-edge of $n_0$ has $r$ as its role, and min greater than or equal to $n$; or ii) $n = 0$.*

4. *$D$ is $(\leq n\, r)$ and some $r$-edge of $n_0$ has $r$ as its role, and max less than or equal to $n$.*

5. *$D$ is $a_1 \cdots a_n \downarrow b_1 \cdots b_m$, and there are rooted paths with label $a_1 \cdots a_n$ and $b_1 \cdots b_m$ in $G$ ending at the same node.*

6. *$D$ is $\forall r.C$, for a role $r$, and either (i) some $r$-edge of $n_0$ has $r$ as its role and $G'$ as its restriction graph with subsumes?$(C, G')$; or (ii) subsumes?$(C, G(\top))$. (Reason: $\forall r.\top \equiv \top$.)*

7. *$D$ is $\forall a.C$, for an attribute $a$, and (i) some $a$-edge of $G$ is of the form $(n_0, a, n')$, and subsumes?$(C, (N, E, n', l))$; or (ii) some $r$-edge of $n_0$ has $a$ as its attribute, and $G'$ as its restriction graph with subsumes?$(C, G')$; or (iii) subsumes?$(C, G(\top))$.*

8. *$D$ is $E \sqcap F$ and both subsumes?$(E, G)$ and subsumes?$(F, G)$ are true.*

There are only two differences between this algorithm and the one for total attributes presented by Borgida and Patel-Schneider (see also Algorithm 2). First, in the partial attribute case, given $D = \forall a.C$, one needs to look up the value restriction either in some $a$-edge or some $r$-edge of $G$, since attributes can label both $a$-edges and $r$-edges. (In the total attribute case, attributes can only label $a$-edges so that examining $r$-edges was not necessary.) The second and most important distinction is the treatment of same-as equalities. As shown in the above algorithm, with $D = a_1 \cdots a_n \downarrow b_1 \cdots b_m$ one only needs to check whether there exist two paths labeled $v := a_1 \cdots a_n$ and $w := b_1 \cdots b_m$ leading the same node in $G$. In the total attribute case, however, it suffices if there exist prefixes $v'$ and $w'$ of $v$ and $w$ with this property, as long as the remaining suffixes are identical.

Soundness and completeness of this algorithm is stated in the following theorem.

**Theorem 3** *Let $C$, $D$ be CLASSIC$^-$ descriptions. Then, $C \sqsubseteq D$ iff subsumes?$(D, G_C)$, where $G_C$ is the canonical form of $G(C)$.*

The soundness of the subsumption algorithm, i.e., the if direction in the theorem stated above, is pretty obvious. As in (Borgida & Patel-Schneider, 1994), the main point of the only-if direction (proof of completeness) is that the canonical graph $G_C$ is deterministic, i.e., from any node, given a role or attribute name $r$, there is at most one outgoing $r$-edge or $a$-edge with $r$ as label. We point the reader to (Borgida & Patel-Schneider, 1994) for the proof, since it is almost identical to the one for total attributes already published there. These proofs reveal that, for the if direction of Theorem 3, description graphs need not be normalized. Thus, one can also show:

**Remark 1** *Let $G$ be some (not necessarily normalized description graph) and let $D$ be a* CLASSIC$^-$ *concept description. Then, subsumes?$(D, G)$ implies $G \sqsubseteq D$.*

Borgida and Patel-Schneider argue that the canonical description graph $G$ of a concept description $C$ can be constructed in time polynomial in the size of $C$. Furthermore, Algorithm 1 runs in time polynomial in the size of $G$ and $D$. It is not hard to see that the changes presented here do not increase the complexity. Thus, soundness and completeness of the subsumption algorithm provides us with the following corollary.

**Corollary 1** *Subsumption for* CLASSIC$^-$ *concept descriptions $C$ and $D$, where attributes are interpreted as partial functions, can be decided in time polynomial in the size of $C$ and $D$.*

## 4. Computing the LCS in CLASSIC$^-$

In this section, we will show that the lcs of two CLASSIC$^-$ concept descriptions can be stated in terms of a product of canonical description graphs. A similar result has been proven by Cohen and Hirsh (1994a) for a sublanguage of CLASSIC$^-$, which only allows for concept names, concept conjunction, value restrictions, and same-as equalities. In particular, this sublanguage does not allow for inconsistent concept descriptions (which, for example, can be expressed by conflicting number-restrictions). Furthermore, the semantics of the description graphs provided by Cohen and Hirsh restricts the results to the case when description graphs are acyclic. This excludes, for example, same-as equalities of the form $\epsilon \downarrow$ spouse $\circ$ spouse.

In the following, we first define the product of description graphs. Then, we show that for given concept descriptions $C$ and $D$, the lcs is equivalent to a description graph obtained as the product of $G_C$ and $G_D$. Our constructions and proofs will be quite close to those in (Cohen & Hirsh, 1994a).

### 4.1 The Product of Description Graphs

A description graph represents the constraints that must be satisfied by all individuals in the extension of the graph. Intuitively, the product of two description graphs is the intersection of these constraints—as the product of finite automata corresponds to the intersection of the words accepted by the automata. However, in the definition of the product of description graphs special care has to be taken of incoherent nodes, i.e., nodes labeled with $\perp$. Also, since attributes may occur both in r-edges and a-edges, one needs to take the product between restriction graphs of r-edges, on the one hand, and the original graphs $G_1$ or $G_2$ (rooted at certain nodes), on the other hand.

**Definition 8** *Let $G_1 = (N_1, E_1, n_1, l_1)$ and $G_2 = (N_2, E_2, n_2, l_2)$ be two description graphs. Then, the product $G := G_1 \times G_2 := (N, E, n_0, l)$ of the two graphs is recursively defined as follows:*

1. $N := N_1 \times N_2$;

2. $n_0 := (n_1, n_2)$;

3. $E := \{((n, n'), a, (m, m')) \mid (n, a, m) \in E_1 \text{ and } (n', a, m') \in E_2\}$;

4. *Let $n \in N_1$ and $n' \in N_2$. If $l_1(n) = \bot$, then let $l((n, n')) := l_2(n')$ and, analogously, if $l_2(n') = \bot$, then $l((n, n')) := l_1(n)$. Otherwise, for $l_1(n) = (S_1, H_1)$ and $l_2(n') = (S_2, H_2)$, define $l((n, n')) := (S, H)$ where*

   (a) *$S := S_1 \cap S_2$;*

   (b) *$H :=$*
     *$\{(r, min(p_1, p_2), max(q_1, q_2), G'_1 \times G'_2) \mid (r, p_1, q_1, G'_1) \in H_1, (r, p_2, q_2, G'_2) \in H_2\} \cup$*
     *$\{(a, 0, 1, G_{1|m} \times G'_2) \mid (n, a, m) \in E_1, (a, p_2, q_2, G'_2) \in H_2\} \cup$*
     *$\{(a, 0, 1, G'_1 \times G_{2|m}) \mid (a, p_1, q_1, G'_1) \in H_1, (n', a, m) \in E_2\}.$*

According to this definition, if in the tuple $(n, n')$ some node, say $n$, is incoherent, then the label of $(n, n')$ coincides with the one for $n'$. The reason for defining the label in this way is that $lcs(\bot, C) \equiv C$ for every concept description $C$. This has been overlooked by Frazier and Pitt (1996), thus making their constructions and proofs only hold for concept descriptions that do not contain inconsistent subexpressions.

Note that $G$, as defined here, might not be connected, i.e., it might contain nodes that cannot be reached from the root $n_0$. Even if $G_1$ and $G_2$ are connected this can happen because all tuples $(n_1, n_2)$ belong to the set of nodes of $G$ regardless of whether they are reachable from the root or not. However, as already mentioned in Section 3.1 we may assume $G$ to be connected.

Also note that the product graph can be translated back into a CLASSIC$^-$ concept description since the product of two description graphs is once again a description graph.

## 4.2 Computing the LCS

We now prove the main theorem of this subsection, which states that the product of two description graphs is equivalent to the lcs of the corresponding concept descriptions.

**Theorem 4** *Let $C_1$ and $C_2$ be two concept descriptions, and let $G_1$ and $G_2$ be corresponding canonical description graphs. Then, $C_{G_1 \times G_2} \equiv lcs(C_1, C_2)$.*

**Proof.** Let $G := G_1 \times G_2$. We will only sketch the proof showing that $C_G$ subsumes $C_1$ and, by symmetry, also $C_2$ (see (Küsters & Borgida, 1999) for details). By construction, if there are two rooted paths to a common node in $G$, then $G_1$ has corresponding paths leading to the same node as well. Thus, by Theorem 3, the same-as equalities in $C_G$ subsume the ones in $C_1$. Now, let $T$ be a spanning tree of $G$, $(m_1, m_2)$ be a node in $G$, and $v$ be the label of the rooted path in $T$ to $(m_1, m_2)$. Then, by construction it follows that there exists a rooted path in $G_1$ to $m_1$ labeled $v$. Furthermore, a rather straightforward inductive proof shows that the concept description $E$ corresponding to the label of $(m_1, m_2)$ subsumes $G_{1|m_1}$. This implies $\forall v.E \sqsupseteq G_1$. As a result, we can conclude $G \sqsupseteq G_1$.

The more interesting part of the proof is to show that $C_G$ is not only a common subsumer of $C_1$ and $C_2$, but the *least* common subsumer.

We now show by induction over the size of $D$, $C_1$, and $C_2$ that if $D$ subsumes $C_1$ and $C_2$, then $D$ subsumes $C_G$: We distinguish different cases according to the definition of "*subsumes?*". Let $G_1 = (N_1, E_1, n_1, l_1)$ be the canonical description graph of $C_1$, $G_2 = (N_2, E_2, n_2, l_2)$ be the canonical description graph of $C_2$, and $G = (N, E, n_0, l) = G_1 \times G_2$. In the following, we assume that $C_1 \sqsubseteq D$ and $C_2 \sqsubseteq D$; thus, *subsumes?*$(D, G_1)$ and $subsumes?(D, G_2)$. We show that $subsumes?(D, G)$. Then, Remark 1 implies $G \sqsubseteq D$, and thus, $C_G \sqsubseteq D$. Note that one cannot use Theorem 3 since $G$ might not be a canonical description graph.

1. If $G$ is incoherent, then there is nothing to show.

2. If $D$ is a concept name, $\top$, or a number-restriction, then by definition of the label of $n_0$ it is easy to see that $subsumes?(D, G)$.

3. If $D$ is $v \downarrow w$, then there exist nodes $m_1$ in $G_1$ and $m_2$ in $G_2$ such that there are two paths from $n_1$ to $m_1$ with label $v$ and $w$, respectively, as well as two paths from $n_2$ to $m_2$ with label $v$ and $w$. Then, by definition of $G$ it is easy to see that there are two paths from $n_0 = (n_1, n_2)$ to $(m_1, m_2)$ with label $v$ and $w$, respectively. This shows $subsumes?(D, G)$.

4. If $D$ is $\forall r.C$, $r$ a role or attribute, then one of several cases applies:

   (i) $n_1$ and $n_2$ have r-edges with role or attribute $r$, and restriction graphs $G'_1$ and $G'_2$, respectively, such that $subsumes?(C, G'_1)$ and $subsumes?(C, G'_2)$;

   (ii) without loss of generality, $n_1$ has an a-edge pointing to $m_1$ with attribute $r$, such that $subsumes?(C, G'_1)$, where $G'_1 := G_{1|m_1}$; and $n_2$ has an r-edge with restriction graph $G'_2$ such that $subsumes?(C, G'_2)$.

   In both cases (i) and (ii), $subsumes?(C, G'_1 \times G'_2)$ follows by induction. Furthermore, by definition of $G$ there is an r-edge with role $r$ and restriction graph $G'_1 \times G'_2$ for $n_0$. This implies $subsumes?(D, G)$.

   (iii) $n_1$ and $n_2$ have a-edges with attribute $r$ leading to nodes $m_1$ and $m_2$, respectively. Then, $subsumes?(C, G_{1|m_1})$ and $subsumes?(C, G_{2|m_2})$. By induction, we know $subsumes?(C, G_{1|m_1} \times G_{2|m_2})$. It is easy to see that $G_{|(m_1, m_2)} = G_{1|m_1} \times G_{2|m_2}$. Furthermore, by definition there is an a-edge with attribute $r$ from $(n_1, n_2)$ to $(m_1, m_2)$ in $G$. This shows $subsumes?(D, G)$.

   (iv) (without loss of generality) $n_1$ has no r-edge and no a-edge with role or attribute $r$. This implies $subsumes?(C, G(\top))$, which also ensures $subsumes?(D, G)$.

5. If $D$ is $E \sqcap F$, then by definition of the subsumption algorithm, $subsumes?(E, G_1)$ and $subsumes?(E, G_2)$ hold. By induction, we have $subsumes?(E, G)$, and analogously, $subsumes?(F, G)$. Thus, $subsumes?(D, G)$. $\qquad\Box$

As stated in Section 3.5, a canonical description graph for a CLASSIC$^-$ concept description can be computed in time polynomial in the size of the concept description. It is not hard to verify that the product of two description graphs can be computed in time polynomial in the size of the graphs. In addition, the concept description corresponding to a description graph can be computed in time polynomial in the size of the graph. Thus, as a consequence of Theorem 4 we obtain:

**Corollary 2** *The lcs of two* CLASSIC$^-$ *concept descriptions always exists and can be computed in time polynomial in the size of the concept descriptions.*

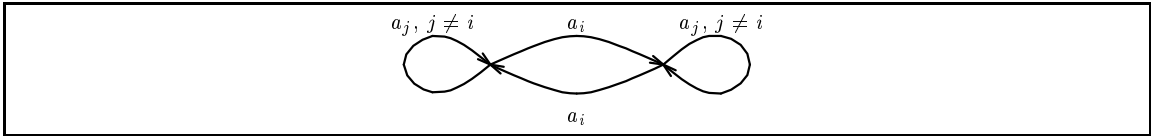

Figure 3: The canonical description graph for $D_i$, without node labels.

As intimated in (Cohen et al., 1992), this statement does not hold for sequences of concept descriptions. Intuitively, generalizing the lcs algorithm to sequences of, say, $n$ concept descriptions, means computing the product of $n$ description graphs. The following proposition shows that the size of such a product graph may grow exponentially in $n$. Thus, the lcs computed in this way grows exponentially in the size of the given sequence. However, this does not imply that this exponential blow-up is unavoidable. There might exist a smaller, still equivalent representation of the lcs. Nevertheless, we can show that the exponential growth is inevitable.

**Proposition 1** *For all integers $n \geq 2$ there exists a sequence $D_1, \ldots, D_n$ of* CLASSIC⁻ *concept descriptions such that the size of every* CLASSIC⁻ *concept description equivalent to $lcs(D_1, \ldots, D_n)$ is at least exponential in $n$ where the size of the $D_i$'s is linear in $n$.*

**Proof.** As in Cohen et al. (1992), for a given $n$, define the concept descriptions $D_i$ as follows:

$$D_i := \prod_{j \neq i} (\varepsilon \downarrow a_j) \sqcap \prod_{j \neq i} (a_i \downarrow a_i a_j) \sqcap (\varepsilon \downarrow a_i a_i)$$

where $a_1, \ldots, a_n$ denote attributes. The canonical description graph for $D_i$ is depicted in Figure 3. Using Algorithm 1 it is easy to see that $D_i \sqsubseteq v \downarrow w$ iff the number of $a_i's$ in $v$ and the number of $a_i's$ in $w$ are equal modulo 2 where $v, w$ are words over $\{a_1, \ldots, a_n\}$. This implies that

$$D_1, \ldots, D_n \sqsubseteq v \downarrow w \quad \text{iff} \quad \begin{array}{l} \text{for all } 1 \leq i \leq n \text{ the number of } a_i's \text{ in } v \text{ and} \\ \text{the number of } a_i's \text{ in } w \text{ are equal modulo 2.} \end{array} \quad (1)$$

Let $s \subseteq \{1, \ldots, n\}$ be a non-empty set. We define $v_s := a_{i_1} \cdots a_{i_k}$ where $i_1 < \cdots < i_k$ are the elements of $s$ and $w_s := a_{i_1}{}^3 a_{i_2}{}^3 \cdots a_{i_k}{}^3$ with $a_j{}^3 := a_j a_j a_j$. Now let $E$ be the lcs of $D_1, \ldots, D_n$, and let $G_E$ be the corresponding canonical description graph with root $n_0$. From (1) we know that $E \sqsubseteq v_s \downarrow w_s$ for every $s \subseteq \{1, \ldots, n\}$. Algorithm 1 implies that the paths from $n_0$ in $G_E$ labeled $v_s$ and $w_s$ exist and that they lead to the same node $q_s$. Assume there are non-empty subsets $s, t$ of $\{1, \ldots, n\}$, $s \neq t$, such that $q_s = q_t$. This would imply $E \sqsubseteq v_s \downarrow v_t$ in contradiction to (1). Thus, $s \neq t$ implies $q_s \neq q_t$. Since there are $2^n - 1$ non-empty subsets of $\{1, \ldots, n\}$, this shows that $G_E$ contains at least $2^n - 1$ nodes. The fact that the size of $G_E$ is linear in the size of $E$ completes the proof. $\qquad\square$

This proposition shows that algorithms computing the lcs of sequences are necessarily worst-case exponential. Conversely, based on the polynomial time algorithm for the binary lcs operation, an exponential time algorithm can easily be specified employing the following identity $lcs(D_1, \ldots, D_n) \equiv lcs(D_n, lcs(D_{n-1}, lcs(\cdots lcs(D_2, D_1) \cdots))$.

**Corollary 3** *The size of the lcs of sequences of* CLASSIC$^-$ *concept descriptions can grow exponentially in the size of the sequences and there exists an exponential time algorithm for computing the lcs.*

## 5. The LCS for Same-as and Total Attributes

In the previous sections, attributes were interpreted as partial functions. In this section, we will present the significant changes in computing the lcs that occur when considering total functions instead of partial functions. More precisely, we will look at a sublanguage $\mathcal{S}$ of CLASSIC$^-$ that only allows for concept conjunction and same-as equalities, but where we have the general assumption that *attributes are interpreted as total functions*.

We restrict our attention to the language $\mathcal{S}$ in order to concentrate on the changes caused by going from partial to total functions. We strongly conjecture, however, that the results represented here can easily be transfered to CLASSIC$^-$ by extending the description graphs for $\mathcal{S}$ as in Section 4.

First, we show that in $\mathcal{S}$ the lcs$_t$ of two concept descriptions does not always exist. Then, we will present a polynomial decision algorithm for the existence of an lcs$_t$ of two concept descriptions. Finally, it will be shown that if the lcs$_t$ of two concept descriptions exists, then it might be exponential in the size of the given concept descriptions and it can be computed in exponential time.

In the sequel, we will simply refer to the lcs$_t$ by lcs. Since throughout the section attributes are always assumed to be total, this does not lead to any confusion.

Once again, it may be useful to keep in mind that for total (though not partial) attributes we have $(u \downarrow v) \sqsubseteq_t (u \circ w \downarrow v \circ w)$ for any $u, w, v \in \mathcal{A}^*$, where $\mathcal{A}^*$ is the set of finite words over $\mathcal{A}$, the finite set of attribute names. Indeed, all the differences between partial and total attributes shown in this section finally trace back to this property.

### 5.1 The Existence of the LCS

In this subsection, we prove that the lcs of two concept descriptions in $\mathcal{S}$ does not always exist. Nevertheless, there is always an infinite representation of the lcs, which will be used in the next subsection to characterize the existence of the lcs.

To accomplish the above, we return to the graph-based characterization of t-subsumption proposed by Borgida and Patel-Schneider (1994), and modified for partial attributes in Section 3. For a concept description $C$, let $G_C$ denote the corresponding canonical description graph, as defined in Section 3.4. Its semantics is specified as in Section 3.1, although now the set of interpretations is restricted to allow attributes to be interpreted as total functions only.

Since $\mathcal{S}$ contains no concept names and does not allow for value-restrictions, the nodes in $G_C$ do not contain concept names and the set of r-edges is empty. Therefore, $G_C$ can be defined by the triple $(N, E, n_0)$ where $N$ is a finite set of nodes, $E$ is a finite set over $N \times \mathcal{A} \times N$, and $n_0$ is the root of the graph.

As a corollary of the results of Borgida and Patel-Schneider, subsumption $C \sqsubseteq_t D$ of concept descriptions $C$ and $D$ in $\mathcal{S}$ can be decided with the following algorithm, which also provides us with a characterization of t-subsumption.

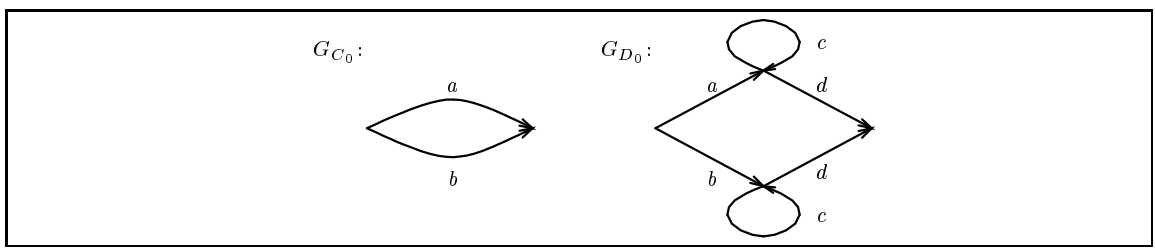

Figure 4: The canonical graphs for $C_0$ and $D_0$

**Algorithm 2** *Let $C$, $D$ be concept descriptions in $\mathcal{S}$, and $G_C = (N, E, n_0)$ be the canonical description graph of $C$. Then, subsumes$_t$?$(D, G_C)$ is defined to be true if and only if one of the following conditions hold:*

1. *$D$ is $v \downarrow w$ and there are words $v', w', u \in \mathcal{A}^*$ such that $v = v'u$ and $w = w'u$, and there are rooted paths in $G_C$ labeled $v'$ and $w'$, respectively, ending at the same node.*

2. *$D$ is $D_1 \sqcap D_2$ and both subsumes$_t$?$(D_1, G_C)$ and subsumes$_t$?$(D_2, G_C)$ are true.*

Apart from the additional constructors handled by Algorithm 1, Algorithm 2 only differs from Algorithm 1 in that, for total attributes, as considered here, it is sufficient if *prefixes* of rooted paths $v$ and $w$ lead to a common node, as long as the remainder in both cases is the same path.

**Theorem 5** *There are concept descriptions in $\mathcal{S}$ such that the lcs of these concept descriptions does not exist in $\mathcal{S}$.*

This result corrects the statement of Cohen et al. (1992) that the lcs always exists, a statement that inadvertently assumed that attributes were partial, not total.

As proof, we offer the following $\mathcal{S}$-concept descriptions, which are shown not to have an lcs:

$$C_0 := a \downarrow b,$$
$$D_0 := a \downarrow ac \sqcap b \downarrow bc \sqcap ad \downarrow bd.$$

The graphs for these concepts are depicted in Figure 4.

The following statement shows that an lcs $E$ of $C_0$ and $D_0$ would satisfy a condition which does not have a "regular structure". This statement can easily be verified using Algorithm 2.

$$E \sqsubseteq_t v \downarrow w \quad \text{iff} \quad v = w \text{ or there exists a nonnegative integer } n \text{ and } u \in$$
$$\mathcal{A}^* \text{ such that } v = ac^n du \text{ and } w = bc^n du \text{ or vice versa.}$$

Given this description of the lcs of $C_0$ and $D_0$, one can show, again, by employing Algorithm 2, that no finite description graph can be equivalent to $E$. However, we omit this elementary proof here, because the absence of the lcs also follows from Theorem 6, where infinite graphs are used to characterize the existence of an lcs. Note that in the partial attribute case, the lcs of $C_0$ and $D_0$ is equivalent to $a \downarrow a \sqcap b \downarrow b$, a result that can be obtained by the lcs algorithm presented in the previous section. The corresponding (finite) description graph consists of a root and two additional nodes, where the root has two outgoing edges leading to the two nodes and labeled $a$ and $b$, respectively.

To state Theorem 6, we first introduce infinite description graphs and show that there always exists an infinite description graph representing the lcs of two $\mathcal{S}$-concept descriptions.

An *infinite description graph* $G$ is defined, like a finite graph, by a triple $(N, E, n_0)$ except that the set of nodes $N$ and the set of edges $E$ may be infinite. As in the finite case, $nvn' \in G$ means that $G$ contains a path from $n$ to $n'$ labeled with the word $v \in \mathcal{A}^*$. The semantics of infinite graphs is defined as in the finite case. Furthermore, infinite graphs are translated into concept descriptions as follows: take an (infinite) spanning tree $T$ of $G$, and, as in the finite case, for every edge of $G$ not contained in it, add to $C_G$ a same-as equality. Note that in contrast to the partial attribute case, $C_G$ need not contain same-as equalities of the form $v \downarrow v$ since, for total attributes, $v \downarrow v \equiv \top$. Still, $C_G$ might be a concept description with an infinite number of conjuncts (thus, an *infinite concept description*). The semantics of such concept descriptions is defined in the obvious way. Analogously to Lemma 2, one can show that an (infinite) graph $G$ and its corresponding (infinite) concept description $C_G$ are equivalent, i.e., $C_G \equiv G$.

We call an (infinite) description graph $G$ *deterministic* if, and only if, for every node $n$ in $G$ and every attribute $a \in \mathcal{A}$ there exists at most one $a$-successor for $n$ in $G$. The graph $G$ is called *complete* if for every node $n$ in $G$ and every attribute $a \in \mathcal{A}$ there is (at least) one $a$-successor for $n$ in $G$. Clearly, for a deterministic and complete (infinite) description graph, every path is uniquely determined by its starting point and its label.

Algorithm 2 (which deals with finite description graphs $G_C$) can be generalized to deterministic and complete (infinite) description graphs $G$ in a straightforward way. To see this, first note that a (finite) description graph coming from an $\mathcal{S}$-concept description is canonical iff it is deterministic in the sense just introduced. Analogously, a deterministic infinite graph can be viewed as being canonical. Thus, requiring (infinite) graphs to be deterministic satisfies the precondition of Algorithm 2. Now, if in addition these graphs are complete, then (unlike the condition stated in the subsumption algorithm) it is no longer necessary to consider prefixes of words because a complete graph contains a rooted path for every word. More precisely, if $v'$ and $w'$ lead to the same node, then this is the case for $v = v'u$ and $w = w'u$ as well, thus making it unnecessary to consider the prefixes $v'$ and $w'$ of $v$ and $w$, respectively. Summing up, we can conclude:

**Corollary 4** *Let* $G = (N, E, n_0)$ *be a deterministic and complete (infinite) description graph and* $v, w \in \mathcal{A}^*$. *Then,*

$$G \sqsubseteq_t v \downarrow w \quad \text{iff} \quad n_0 vn \in G \text{ and } n_0 wn \in G \text{ for some node } n.$$

We shall construct an (infinite) graph representing the lcs of two concept descriptions in $\mathcal{S}$ as the product of the so-called completed canonical graphs. This infinite representation of the lcs will be used later to characterize the existence of an lcs in $\mathcal{S}$, i.e., the existence of a finite representation of the lcs.

We now define the completion of a graph. Intuitively, a graph is completed by iteratively adding outgoing a-edges labeled with an attribute $a$ for every node in the graph that does not have such an outgoing a-edge. This process might extend a graph by infinite trees. As an example, the completion of $G_{C_0}$ (cf. Figure 4) is depicted in Figure 5 with $\mathcal{A} = \{a, b, c, d\}$.

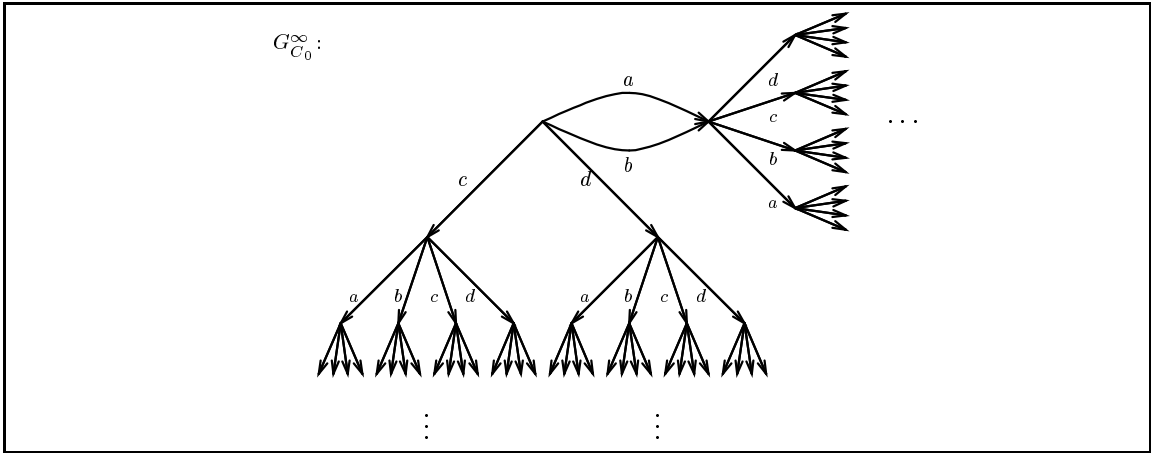

Figure 5: The complete graph for $C_0$

Formally, completions are defined as follows: Let $G$ be an (infinite) description graph. The graph $G'$ is an *extension* of $G$ if for every node $n$ in $G$ and for every attribute $a \in \mathcal{A}$ such that $n$ has no outgoing edges labeled $a$, a new node $m_{n,a}$ is added, as well as an edge $(n, a, m_{n,a})$. Now, let $G^0, G^1, G^2, \ldots$ be a sequence of graphs such that $G^0 = G$ and $G^{i+1}$ is an extension of $G^i$, for $i \geq 0$. If $G^i = (N_i, E_i, n_0)$, then

$$G^\infty := (\bigcup_{i \geq 0} N_i, \bigcup_{i \geq 0} E_i, n_0)$$

is called the *completion* of $G$. By construction, $G^\infty$ is a complete graph. Furthermore, if $G$ is deterministic, then $G^\infty$ is deterministic as well. Finally, it is easy to see that a graph and its extension are equivalent. Thus, by induction, $G^\infty \equiv_t G$.

The nodes in $\bigcup_{i \geq 1} N_i$, i.e., the nodes in $G^\infty$ that are not in $G$, are called *tree nodes*; the nodes of $G$ are called *non-tree nodes*. By construction, for every tree node $t$ in $G^\infty$ there is exactly one direct predecessor of $t$ in $G^\infty$, i.e., there is exactly one node $n$ and one attribute $a$ such that $(n, a, t)$ is an edge in $G^\infty$; $n$ is called *a-predecessor* of $t$. Furthermore, there is exactly one youngest ancestor $n$ in $G$ of a tree node $t$ in $G^\infty$; $n$ is the *youngest ancestor* of $t$ if there is a path from $n$ to $t$ in $G^\infty$ which does not contain non-tree nodes except for $n$. Note that there is only one path from $n$ to $t$ in $G^\infty$. Finally, observe that non-tree nodes have only non-tree nodes as ancestors.

Note that the completion of a canonical description graph is always complete and deterministic.

In the sequel, let $C$, $D$ be two concept descriptions in $\mathcal{S}$, $G_C = (N_C, E_C, n_C)$, $G_D = (N_D, E_D, n_D)$ be their corresponding canonical graphs, and $G_C^\infty$, $G_D^\infty$ be the completions of $G_C$, $G_D$. The products $G := G_C \times G_D$ and $G_\infty^\times := G_C^\infty \times G_D^\infty$ are specified as in Definition 1. As usual, we may assume $G$ and $G_\infty^\times$ are connected, i.e., they only contain nodes that are reachable from the root $(n_C, n_D)$; otherwise, one can remove all those nodes that cannot be reached from the root without changing the semantics of the graphs.

We denote the product $G_C^\infty \times G_D^\infty$ by $G_\infty^\times$ instead of $G^\infty$ (or $G_\times^\infty$) because otherwise this graph could be confused with the completion of $G$. In general, these graphs do not coincide. As an example, take the products $G_{C_0} \times G_{D_0}$ and $G_{C_0}^\infty \times G_{D_0}^\infty$ (see Figure 4 for the graphs $G_{C_0}$ and $G_{D_0}$). The former product results in a graph that consists of a root with two outgoing a-edges, one labeled $a$ and the other one labeled $b$. (As mentioned before, this graph corresponds to the lcs of $C_0$ and $D_0$ in the partial attribute case.) The product of the completed graphs, on the other hand, is a graph that is obtained as the completion of the graph depicted in Figure 6 (the infinite trees are omitted for the sake of simplicity).

As an easy consequence of the fact that $G_C \equiv G_C^\infty$ and Corollary 4, one can prove the following lemma.

**Lemma 6** $C \sqsubseteq_t v \downarrow w$ iff $n_C v n \in G_C^\infty$ and $n_C w n \in G_C^\infty$ for a node $n$ in $G_C^\infty$.

But then, by the construction of $G_\infty^\times$ we know:

**Proposition 2** $C \sqsubseteq_t v \downarrow w$ and $D \sqsubseteq_t v \downarrow w$ iff $(n_C, n_D) v n \in G_\infty^\times$ and $(n_C, n_D) w n \in G_\infty^\times$ for a node $n$ in $G_\infty^\times$.

In particular, $G_\infty^\times$ represents the lcs of the concept descriptions $C$ and $D$ in the following sense:

**Corollary 5** The (infinite) concept description $C_{G_\infty^\times}$ corresponding to $G_\infty^\times$ is the lcs of $C$ and $D$, i.e., i) $C, D \sqsubseteq_t C_{G_\infty^\times}$ and ii) $C, D \sqsubseteq_t E'$ implies $C_{G_\infty^\times} \sqsubseteq_t E'$ for every $\mathcal{S}$-concept description $E'$.

## 5.2 Characterizing the Existence of an LCS

Let $C$, $D$ be concept descriptions in $\mathcal{S}$ and let the graphs $G_C$, $G_D$, $G$, $G_C^\infty$, $G_D^\infty$, and $G_\infty^\times$ be defined as above.

We will show that $G_\infty^\times$ not only represents a (possibly infinite) lcs of the $\mathcal{S}$-concept descriptions $C$ and $D$ (Corollary 5), but that $G_\infty^\times$ can be used to characterize the existence of a finite lcs. The existence depends on whether $G_\infty^\times$ contains a finite or an infinite number of so-called same-as nodes.

**Definition 9** A node $n$ of an (infinite) description graph $H$ is called a same-as node if there exist two direct predecessors of $n$ in $H$. (The a-edges leading to $n$ from these nodes may be labeled differently.)

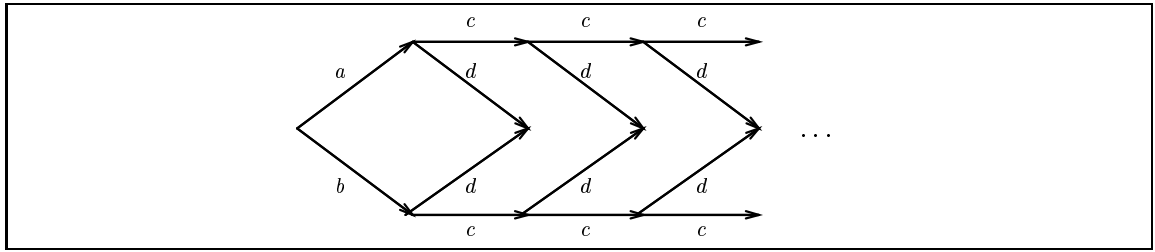

Figure 6: A subgraph of $G_{C_0}^\infty \times G_{D_0}^\infty$

For example, the graph depicted in Figure 6 contains an infinite number of same-as nodes. We will show that this is a sufficient and necessary condition for the lcs of $C_0$ and $D_0$ not to exist.

It is helpful to observe that same-as nodes in $G_\infty^\times$ have one of the forms $(g, f)$, $(f, t)$, and $(t, f)$, where $g$ and $f$ are non-tree nodes and $t$ is a tree node. There cannot exist a same-as node of the form $(t_1, t_2)$, where both $t_1$ and $t_2$ are tree nodes, since tree nodes only have exactly one direct predecessor, and thus $(t_1, t_2)$ does. Moreover, if $G_\infty^\times$ has an infinite number of same-as nodes, it must have an infinite number of same-as nodes of the form $(f, t)$ or $(t, f)$, because there only exist a finite number of nodes in $G_\infty^\times$ of the form $(g, f)$. For this reason, in the following lemma we only characterize same-as nodes of the form $(f, t)$. (Nodes of the form $(t, f)$ can be dealt with analogously.) To state the lemma, recall that with $n_0 u \cdot n_1 \cdot v n_2 \in H$, for some graph $H$, we describe a path in $H$ labeled $uv$ from $n_0$ to $n_2$ that passes through node $n_1$ after $u$ (i.e., $n_0 u n_1 \in H$ and $n_1 v n_2 \in H$); this is generalized the obvious way to interpret $n_0 u_1 \cdot n_1 \cdot u_2 \cdot n_2 \cdot u_3 n_3 \in H$.

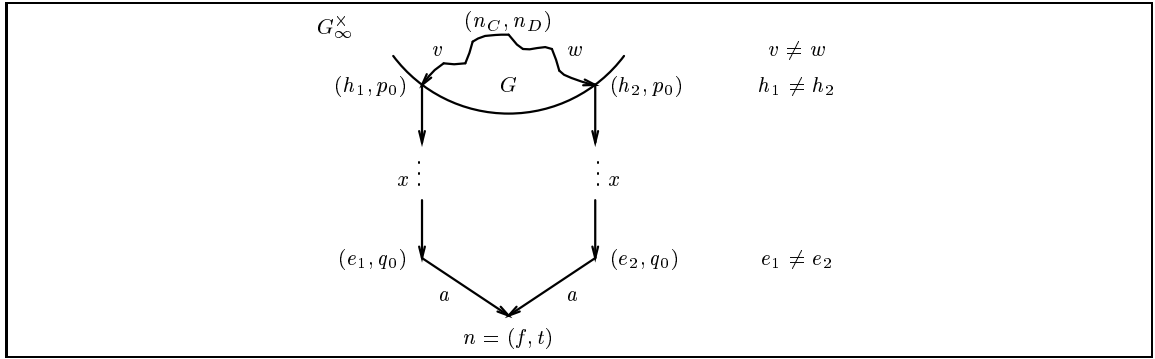

Figure 7: same-as nodes in $G_\infty^\times$

**Lemma 7** *Given a node $f$ in $G_C$ and a tree-node $t$ in $G_D^\infty$, the node $n = (f, t)$ in $G_\infty^\times$ is a same-as node iff*

- *there exist nodes $(h_1, p_0)$, $(h_2, p_0)$ in $G$, $h_1 \neq h_2$;*

- *there exist nodes $(e_1, q_0)$, $(e_2, q_0)$ in $G_\infty^\times$, where $e_1$, $e_2$ are distinct nodes in $G_C$ and $q_0$ is a node in $G_D^\infty$; and*

- *there exists an attribute $a \in \mathcal{A}$ and $v, w, x \in \mathcal{A}^*$, $v \neq w$, where $\mathcal{A}$ is the set of attributes in $C$,*

*such that*

$$(n_C, n_D) v \cdot (h_1, p_0) \cdot x \cdot (e_1, q_0) \cdot a (f, t) \text{ and } (n_C, n_D) w \cdot (h_2, p_0) \cdot x \cdot (e_2, q_0) \cdot a (f, t)$$

*are paths in $G_\infty^\times$ (see Figure 7). For the direct successors $(h_1', p_0')$ and $(h_2', p_0')$ of $(h_1, p_0)$ and $(h_2, p_0)$ in this paths, we, in addition, require $p_0'$ to be a tree node in $G_D^\infty$.[4]*

---

4. Note that since $G_\infty^\times$ is deterministic, the successors of $(h_1, p_0)$ and $(h_2, p_0)$ in the two paths must in fact be of the form $(\cdot, p_0')$.

**Proof.** The if direction is obvious. We proceed with the only-if direction and assume that $n$ is a same-as node in $G_\infty^\times$. Let $p_0$ be the (uniquely determined) youngest ancestor of $t$ in $G_D^\infty$. In particular, $p_0$ is a node in $G_D$ and there exists $p_0 x \cdot q_0 \cdot at$ in $G_D^\infty$ with $a \in \mathcal{A}$ and $x \in \mathcal{A}^*$ such that the successor of $p_0$ in this path is a tree node in $G_D$.

Since $n$ is a same-as node and $t$ can only be reached via $q_0$ and the attribute $a$, there must exist $e_1$, $e_2$ in $G_C$, $e_1 \neq e_2$, with $(e_1, q_0) a(f, t), (e_2, q_0) a(f, t) \in G_\infty^\times$. Since $G_\infty^\times$ is connected, there are paths from $(n_C, n_D)$ to $(e_1, q_0)$ and $(e_2, q_0)$. Every path from $n_D$ to $q_0$ must pass through $p_0$ and the suffix of the label of this path is $x$. Consequently, there exist nodes $h_1, h_2$ in $G_C$ such that $(h_1, p_0) x \cdot (e_1, q_0) \cdot a(f, t)$ and $(h_2, p_0) x \cdot (e_2, q_0) \cdot a(f, t)$ are paths in $G_\infty^\times$. In particular, $xa$ is a label of a path from $h_1$ to $f$ in $G_C$, and the label $xa$ only consists of attributes contained in $C$. If $h_1 = h_2$, then this, together with the fact that $G_C$ is deterministic, would imply $e_1 = e_2$. Hence, $h_1 \neq h_2$. Let $v$, $w$ be the labels of the paths from $(n_C, n_D)$ to $(h_1, p_0)$ and $(h_2, p_0)$, respectively. As $G$ is deterministic and $h_1 \neq h_2$, it follows that $v \neq w$. $\qquad\square$

The main results of this section is stated in the next theorem. As a direct consequence of this theorem, we obtain that there exists no lcs in $\mathcal{S}$ for the concept descriptions $C_0$ and $D_0$ of our example.

**Theorem 6** *The lcs of $C$ and $D$ exists iff the number of same-as nodes in $G_\infty^\times$ is finite.*

**Proof.** We start by proving the only-if direction. For this purpose, we assume that $G_\infty^\times$ contains an infinite number of same-as nodes and show that there is no (finite) lcs for $C$ and $D$ in $\mathcal{S}$.

As argued before, we may assume that $G_\infty^\times$ contains an infinite number of same-as nodes of the form $(f, t)$ or $(t, f)$, where $t$ is a tree node and $f$ is a non-tree node. More precisely, say $G_\infty^\times$ contains for every $i \geq 1$ nodes $n_i = (f_i, t_i)$ such that $f_i$ is a node in $G_C$ and $t_i$ is a tree node in $G_D^\infty$. According to Lemma 7, for every same-as node $n_i$ there exist nodes $h_{1,i}, h_{2,i}, e_{1,i}, e_{2,i}$ in $G_C$, $p_{0,i}$ in $G_D$, and $q_{0,i}$ in $G_D^\infty$ as well as $a_i \in \mathcal{A}$ and $x_i \in \mathcal{A}^*$ with the properties required in Lemma 7.

Since $G_C$ and $G_D$ are finite description graphs, the number of tuples of the form $h_{1,i}, h_{2,i}, e_{1,i}, e_{2,i}, f_i, a_i$ is finite. Thus, there must be an infinite number of $i$'s yielding the same tuple $h_1, h_2, e_1, e_2, f, a$. In particular, $h_1 \neq h_2$ and $e_1 \neq e_2$ are nodes in $G_C$ and there is an infinite number of same-as nodes of the form $n_i = (f, t_{1,i})$. Finally, as in the lemma, let $v$, $w$ be the label of paths (in $G$) from $(n_C, n_D)$ to $(h_1, p_0)$ and $(h_2, p_0)$.

Now, assume there is an lcs $E$ of $C$ and $D$ in $\mathcal{S}$. According to Corollary 5, $E \equiv_t C_{G_\infty^\times}$. Let $G_E$ be the finite canonical graph for $E$ with root $n'$. By Proposition 2 and Lemma 7 we know $E \sqsubseteq_t vx_i a \downarrow wx_i a$. From Algorithm 2 it follows that there are words $v'$, $w'$, and $u$ such that $vx_i a = v'u$ and $wx_i a = w'u$, where the paths in $G_E$ starting from $n'$ labeled $v'$, $w'$ lead to the same node in $G_E$.

If $u \neq \varepsilon$, then $u = u'a$ for some word $u'$. Then, Algorithm 2 ensures $E \sqsubseteq_t vx_i \downarrow wx_i$. However, by Lemma 7 we know that the words $vx_i$ and $wx_i$ lead to different nodes in $G_\infty^\times$, namely, $(e_1, q_{0,i})$ and $(e_2, q_{0,i})$, which, with Proposition 2, leads to the contradiction $E \equiv G_\infty^\times \not\sqsubseteq_t vx_i \downarrow wx_i$. Thus, $u = \varepsilon$.

As a result, for every $i \geq 1$ there exists a node $q_i$ in $G_E$ such that $n'vx_i aq_i$ and $n'wx_i aq_i$ are paths in $G_E$. Because $G_E$ is a finite description graph, there exist $i, j \geq 1$, $i \neq j$, with $q_i = q_j$. By Algorithm 2, this implies $E \sqsubseteq_t vx_i a \downarrow wx_j a$. On the other hand, the path in $G_\infty^\times$ starting from $(n_C, n_D)$ with label $vx_i a$ leads to the node $n_i$ and the one for $wx_j a$ leads to $n_j$. Since $n_i \neq n_j$, Proposition 2 implies $E \not\sqsubseteq_t vx_i a \downarrow wx_j a$, which is a contradiction. To sum up, we have shown that there does not exist an lcs for $C$ and $D$ in $\mathcal{S}$.

This shows that there is no lcs of $C$, $D$ in $\mathcal{S}$ which completes the proof of the only-if direction.

We now prove the if direction of Theorem 6. For this purpose, we assume that $G_\infty^\times$ has only a finite number of same-as nodes. Note that every same-as node in $G_\infty^\times$ has only a finite number of direct predecessors. To see this, two cases are distinguished: i) a node of the form $(g_1, g_2)$ in $G$ has only predecessors in $G$; ii) if $t$ is a tree node and $g$ a non-tree node, then a predecessor of $(g, t)$ in $G_\infty^\times$ is of the form $(g', t')$ where $t'$ is the unique predecessor (tree or non-tree node) of $t$ and $g'$ is a non-tree node. Since the number of nodes in $G_C$ and $G_D$ is finite, in both cases we only have a finite number of predecessors. But then, the spanning tree $T$ of $G_\infty^\times$ coincides with $G_\infty^\times$ except for a finite number of edges because, if $T$ does not contain a certain edge, then this edge leads to a same-as node. As a result, $C_{G_\infty^\times}$ is an $\mathcal{S}$-concept description because it is a finite conjunction of same-as equalities. Finally, Corollary 5 shows that $C_{G_\infty^\times}$ is the lcs of $C$ and $D$. □

If $v \downarrow w$ is a conjunct in $C_{G_\infty^\times}$, then $v$ and $w$ lead from the root of $G_\infty^\times$ to a same-as node. As mentioned before, same-as nodes are of the form $(f, g)$, $(f, t)$, or $(t, f)$, where $t$ is a tree node and $f, g$ are non-tree nodes. Consequently, $v$ and $w$ must be paths in $G_C$ or $G_D$. Thus, they only contain attributes occurring in $C$ or $D$.

**Corollary 6** *If the lcs of two concept description $C$ and $D$ in $\mathcal{S}$ exists, then there is a concept description in $\mathcal{S}$ only containing attributes occurring in $C$ or $D$ that is equivalent to the lcs.*

Therefore, when asking for the existence of an lcs, we can w.o.l.g. assume that the set of attributes $\mathcal{A}$ is finite. This fact will be used in the following two subsections.

### 5.3 Deciding the Existence of an LCS

From the following corollary we will derive the desired decision algorithm for the existence of an lcs of two concept descriptions in $\mathcal{S}$. To state the corollary we need to introduce the language $L_{G_C}(q_1, q_2) := \{w \in \mathcal{A}^* \mid$ there is a path from the node $q_1$ to $q_2$ in $G_C$ labeled $w\}$. Since description graphs can be viewed as finite automata, such a language will be regular. Moreover, let $a\mathcal{A}^*$ denote the set $\{aw \mid w \in \mathcal{A}^*\}$ for an attribute $a \in \mathcal{A}$, where $\mathcal{A}$ is a finite alphabet.

**Corollary 7** *$G_\infty^\times$ contains an infinite number of same-as nodes iff either*
*(i) there exist nodes $(h_1, p_0)$, $(h_2, p_0)$ in $G$ as well as nodes $f, e_1, e_2$ in $G_C$, and attributes $a, b \in \mathcal{A}$ such that*

1. *$h_1 \neq h_2$, $e_1 \neq e_2$;*

2. *$p_0$ does not have a $b$-successor in $G_D$;*

3. *$(e_1, a, f)$, $(e_2, a, f)$ are edges in $G_C$; and*

4. $L_{G_C}(h_1, e_1) \cap L_{G_C}(h_2, e_2) \cap b\mathcal{A}^*$ *is an infinite set of words;*

*or*

*(ii) the same statement as (i) but with rôles of $C$ and $D$ switched.*

**Proof.** We first prove the only-if direction. Assume that $G_\infty^\times$ contains an infinite number of same-as nodes. Then, w.l.o.g., we find the configuration in $G_\infty^\times$ described in the proof of Theorem 6. This configuration satisfies the conditions 1. and 3. stated in the corollary. If, for $i \neq j$, the words $x_i$ and $x_j$ coincide, we can conclude $n_i = n_j$ because $G_\infty^\times$ is a deterministic graph. However, by definition, $n_i \neq n_j$. Hence, $x_i \neq x_j$. Because $\mathcal{A}$ is finite, we can, w.l.o.g., assume that all $x_i$'s have $b \in \mathcal{A}$ as their first letter for some fixed $b$. Thus, condition 4. is satisfied as well. According to the configuration, the $b$-successor of $(\cdot, p_0)$ in $G_\infty^\times$ is of the form $(\cdot, p_0')$ where $p_0'$ is a tree node. Thus, $p_0$ does not have a $b$-successor in $G_D$, which means that condition 3. is satisfied.

We now prove the if direction of the corollary. For this purpose, let $bx \in L_{G_C}(h_1, e_1) \cap L_{G_C}(h_2, e_2) \cap b\mathcal{A}^*$. Since $p_0$ has no $b$-successor in $G_D$ it follows that there are tree nodes $t, t'$ in $G_D^\infty$ such that $p_0 bx \cdot t \cdot at' \in G_D^\infty$. Thus, we have $(h_1, p_0) bx \cdot (e_1, t) \cdot a(f, t') \in G_\infty^\times$ and $(h_2, p_0) bx \cdot (e_2, t) \cdot a(f, t') \in G_\infty^\times$. Since $e_1 \neq e_2$, we can conclude $(e_1, t) \neq (e_2, t)$. This means that $(f, t')$ is a same-as node. Analogously, for $by \in L_{G_C}(h_1, e_1) \cap L_{G_C}(h_2, e_2) \cap b\mathcal{A}^*$ there are tree nodes $s, s'$ in $G_D^\infty$ such that $p_0 by \cdot s \cdot as' \in G_D^\infty$ and $(f, s')$ is a same-as node in $G_\infty^\times$. Since $bx$ and $by$ both start with $b$, and the $b$-successor of $p_0$ in $G_D^\infty$ is a tree node, $x \neq y$ implies $s' \neq t'$. Hence, $(f, t')$ and $(f, s')$ are distinct same-as nodes. This shows that if the set $L_{G_C}(h_1, e_1) \cap L_{G_C}(h_2, e_2) \cap b\mathcal{A}^*$ is infinite, $G_\infty^\times$ must have an infinite number of same-as nodes. □

For given nodes $(h_1, p_0)$, $(h_2, p_0)$ in $G$, attributes $a, b \in \mathcal{A}$, nodes $f, e_1, e_2 \in G_C$ the conditions 1. to 3. in Corollary 7 can obviously be checked in time polynomial in the size of the concept descriptions $C$ and $D$. As for the last condition, note that an automaton accepting the language $L_{G_C}(h_1, e_1) \cap L_{G_C}(h_2, e_2) \cap b\mathcal{A}^*$ can be constructed in time polynomial in the size of $C$. Furthermore, for a given finite automaton it is decidable in time polynomial in the size of the automaton if it accepts an infinite language (see the book by Hopcroft and Ullman (1979) for details). Thus, condition 4. can be tested in time polynomial in the size of $C$ and $D$ as well. Finally, since the size of $G$ and $G_C$ is polynomial in the size of $C$ and $D$, only a polynomial number of configurations need to be tested. Together with Corollary 7 these complexities provide us with the following corollary.

**Corollary 8** *For given concept descriptions $C$ and $D$ in $\mathcal{S}$ it is decidable in time polynomial in the size of $C$ and $D$ whether lcs of $C$ and $D$ exists in $\mathcal{S}$.*

## 5.4 Computing the LCS

In this subsection, we first show that the size of an lcs of two $\mathcal{S}$-concept descriptions may grow exponentially in the size of the concept descriptions. This is a stronger result than that presented for partial attributes, where it was only shown that the lcs of a sequence of concept descriptions in $\mathcal{S}$ can grow exponentially. Then, we present an exponential time lcs algorithm for $\mathcal{S}$-concept descriptions.

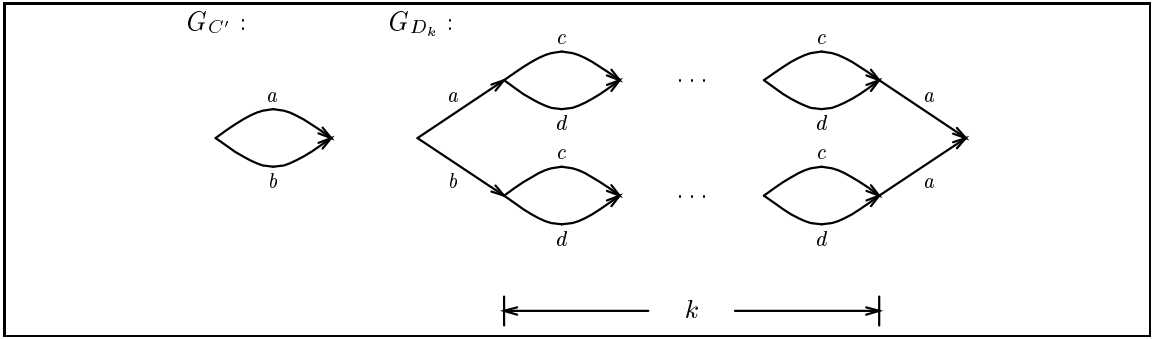

Figure 8: The canonical description graphs for $C'$ and $D_k$

In order to show that the lcs may be of exponential size, we consider the following example, where $\mathcal{A} := \{a, b, c, d\}$. We define

$$C' := a \downarrow b,$$
$$D_k := \prod_{i=1}^{k} ac^i \downarrow ad^i \sqcap \prod_{i=1}^{k} bc^i \downarrow bd^i \sqcap ac^k a \downarrow bc^k a.$$

The corresponding canonical description graphs $G_{C'}$ and $G_{D_k}$ are depicted in Figure 8.

A finite graph representing the lcs of $C'$ and $D_k$ is depicted in Figure 9 for $k = 2$. This graph can easily be derived from $G_{C'}^{\infty} \times G_{D_k}^{\infty}$. The graph comprises two binary trees of height $k$, and thus, it contains at least $2^k$ nodes. In the following, we will show that there is no canonical description graph $G_{E_k}$ (with root $n_0$) representing the lcs $E_k$ of $C'$ and $D_k$ with less than $2^k$ nodes. Let $x \in \{c, d\}^k$ be a word of length $k$ over $\{c, d\}$, and let $v := axa$, $w := bxa$. Using the canonical description graphs $G_{C'}$ and $G_{D_k}$ it is easy to see that $C' \sqsubseteq_t v \downarrow w$ and $D_k \sqsubseteq_t v \downarrow w$. Thus, $E_k \sqsubseteq_t v \downarrow w$. By Algorithm 2, this means that there are words $v', w', u$ such that $v = v'u$, $w = w'u$, and there are paths from $n_0$ labeled $v'$ and $w'$ in $G_{E_k}$ leading to the same node in $G_{E_k}$. Suppose $u \neq \varepsilon$. Then, Algorithm 2 implies $E_k \sqsubseteq_t ax \downarrow bx$. But according to $G_D$, $D \not\sqsubseteq_t ax \downarrow bx$. Therefore $u$ must be the empty word $\varepsilon$. This proves that in $G_{E_k}$ there is a path from $n_0$ labeled $axa$ for every $x \in \{c,d\}^k$. Hence, there is a path for every $ax$. Now, let $y \in \{c,d\}^k$ be such that $x \neq y$. If the paths for $ax$ and $ay$ from $n_0$ in $G_{E_k}$ lead to the same node, then this implies $E_k \sqsubseteq_t ax \downarrow ay$ in contradiction to $C' \not\sqsubseteq_t ax \downarrow ay$. As a result, $ax$ and $ay$ lead to different nodes in $G_{E_k}$. Since $\{c,d\}^k$ contains $2^k$ words, this shows that $G_{E_k}$ has at least $2^k$ nodes. Finally, taking into account that the size of a canonical description graph of a concept description in $\mathcal{S}$ is linear in the size of the corresponding description we obtain the following theorem.

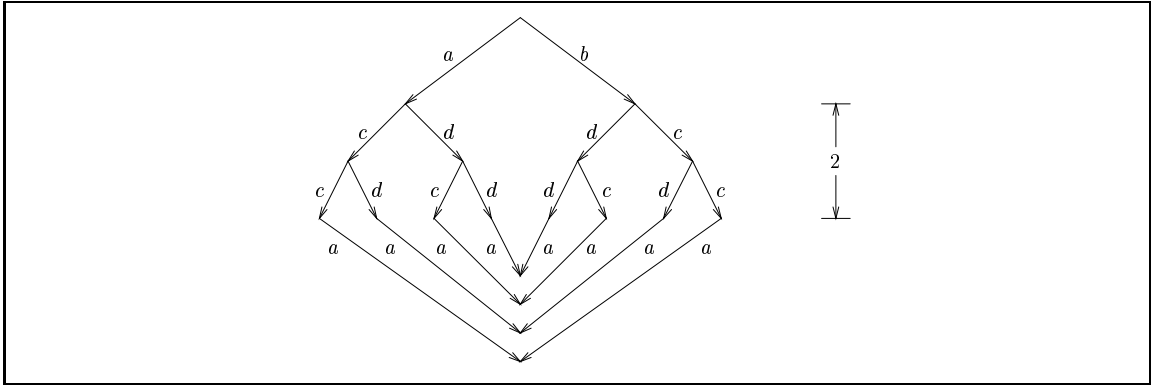

Figure 9: A finite graph representing the lcs of $C'$ and $D_2$

**Theorem 7** *The lcs of two $\mathcal{S}$-concept descriptions may grow exponentially in the size of the concepts.*

The following (exponential time) algorithm computes the lcs of two $\mathcal{S}$-concept descriptions in case it exists.

**Algorithm 3**

**Input:** *concept descriptions $C$, $D$ in $\mathcal{S}$, for which the lcs exists in $\mathcal{S}$;*

**Output:** *lcs of $C$ and $D$ in $\mathcal{S}$;*

1. *Compute $G' := G_C \times G_D$;*

2. *For every combination*

   - *of nodes $(h_1, p_0)$, $(h_2, p_0)$ in $G = G_C \times G_D$, $h_1 \neq h_2$;*
   - *$a \in \mathcal{A}$, $e_1, e_2, f$ in $G_C$, $e_1 \neq e_2$, where $(e_1, a, f)$ and $(e_2, a, f)$ are edges in $G_C$*

   *extend $G'$ as follows: Let $G_{h_1,t}$, $G_{h_2,t}$ be two trees representing the (finite) set of words in*

   $$L := \left( L_{G_C}(h_1, e_1) \cap L_{G_C}(h_2, e_2) \cap \bigcup_{b \notin succ(p_0)} b\mathcal{A}^* \right) \cup \left\{ \begin{array}{ll} \{\varepsilon\}, & \text{if } a \notin succ(p_0) \\ \emptyset, & \text{otherwise} \end{array} \right.$$

   *where $succ(p_0) := \{b \mid p_0 \text{ has a } b\text{-successor}\}$ and the set of nodes of $G_{h_1,t}$, $G_{h_2,t}$, and $G'$ are assumed to be disjoint. Now, replace the root of $G_{h_1,t}$ by $(h_1, p_0)$, the root of $G_{h_2,t}$ by $(h_2, p_0)$, and extend $G'$ by the nodes and edges of these two trees. Finally, add a new node $n_v$ for every word $v$ in $L$, and for each node of the trees $G_{h_1,t}$ and $G_{h_2,t}$ reachable from the root of $G_{h_1,t}$ and $G_{h_2,t}$ by a path labeled $v$, add an edge with label $a$ from it to $n_v$. The extension is illustrated in Figure 10.*

3. *The same as in step 2, with rôles of $C$ and $D$ switched.*

4. *Compute the canonical graph of $G'$, which is called $G'$ again. Then, output the concept description $C_{G'}$ of $G'$.*

**Proposition 3** *The translation $C_{G'}$ of the graph $G'$ computed by Algorithm 3 is the lcs $E$ of $C$ and $D$.*

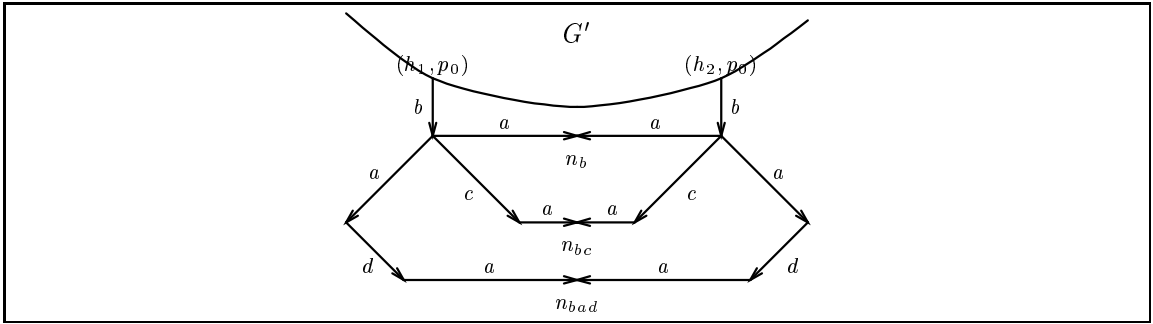

Figure 10: The extension at the nodes $(h_1, p_0)$, $(h_2, p_0)$ in $G'$ where $L = \{b, bc, bad\}$

**Proof.** It is easy to see that if there are two paths in $G'$ labeled $y_1$ and $y_2$ leading from the root $(n_C, n_D)$ to the same node, then $G_\infty^\times$ contains such paths as well. Consequently, $(E \equiv_t) G_\infty^\times \sqsubseteq_t G'$.

Now, assume $E \sqsubseteq_t y_1 \downarrow y_2$, $y_1 \neq y_2$. By Proposition 2 we know that there are paths in $G_\infty^\times$ labeled $y_1$ and $y_2$ leading to the same node $n$. W.l.o.g, we may assume that $n$ is a same-as node in $G_\infty^\times$. Otherwise, there exist words $y_1', y_2', u$ with $y_1 = y_1'u$, $y_2 = y_2'u$ such that $y_1'$ and $y_2'$ lead to a same-as node. If we can show that $G'$ contains paths labeled $y_1'$ and $y_2'$ leading to the same node, then, by Algorithm 2, this is sufficient for $G' \sqsubseteq_t y_1 \downarrow y_2$. So let $n$ be a same-as node. We distinguish two cases:

1. If $n$ is a node in $G = G_C \times G_D$, then the paths for $y_1$ and $y_2$ are paths in $G$. Since $G$ is a subgraph of $G'$ this holds for $G'$ as well. Hence, $C_{G'} \sqsubseteq_t y_1 \downarrow y_2$.

2. Assume $n$ is not a node in $G$. Then, since $n$ is a same-as node, we know that $n$ is of the form $(f, t)$ or $(t, f)$ where $f$ is a non-tree node and $t$ is a tree node. By symmetry, we may assume that $n = (f, t)$. Now it is easy to see that there exist nodes $h_1, h_2, e_1, e_2$ in $G_C$, $p_0$ in $G_D$, and a tree node $q_0$ in $G_D^\infty$ as well as $a \in \mathcal{A}$ and $x, v, w \in \mathcal{A}^*$ as specified in Lemma 7 such that $y_1 = vxa$ and $y_2 = wxa$. But then, with $h_1, h_2, e_1, e_2, p_0, f$ and $a$ the preconditions of Algorithm 3 are satisfied and $x \in L$. Therefore, by construction of $G'$ there are paths labeled $y_1$ and $y_2$, respectively, leading from the root to the same node. $\square$

We note that the product $G$ of $G_C$ and $G_D$ can be computed in time polynomial in the size of $C$ and $D$. Furthermore, there is only a polynomial number of combinations of nodes $(h_1, p_0)$, $(h_2, p_0)$ in $G$, $e_1, e_2, f$ in $G_C$, $a \in \mathcal{A}$. Finally, the finite automaton for $L$ can be computed in time polynomial in the size of $C$ and $D$. In particular, the set of states of this automaton can polynomially be bounded in the size of $C$ and $D$. If $L$ contained a word longer than the number of states, the accepting path in the automaton contains a cycle. But then, the automaton would accept infinitely many words, in contradiction to the assumption that $L$ is finite. Thus, the length of all words in $L$ can be bounded polynomially in the size of $C$ and $D$. In particular, this means that $L$ contains only an exponential number of words. Trees representing these words can be computed in time exponential in the size of $C$ and $D$.

**Corollary 9** *If the lcs of two $\mathcal{S}$-concept descriptions exists, then it can be computed in time exponential in the size of the concept descriptions.*

## 6. Conclusion

Attributes — binary relations that can have at most one value – have been distinguished in many knowledge representation schemes and other object-centered modeling languages. This had been done to facilitate modeling and, in description logics, to help identify tractable sets of concept constructors (e.g., restricting same-as to attributes). In fact, same-as restrictions are quite important from a practical point of view, because they support the modeling of actions and their components (Borgida & Devanbu, 1999).

A second distinction, between attributes as total versus partial functions, had not been considered so essential until now. This paper has shown that this distinction can sometime have significant effects.

In particular, we have first shown that the approach for computing subsumption of CLASSIC concepts with total attributes, presented by Borgida and Patel-Schneider (1994), can be modified to accommodate partial attributes, by treating partial attributes as roles until they participate in same-as restrictions, in which case they are "converted" to total attributes. As a result, we obtain polynomial-time algorithms for subsumption and consistency checking in this case also.

In the case of computing least common subsumers, which was introduced as a technique for learning non-propositional descriptions of concepts, we first noted that several of the papers in the literature (Cohen & Hirsh, 1994a; Frazier & Pitt, 1996) (implicitly) used partial attributes, when considering CLASSIC. Furthermore, these papers used a weaker version of the "concept graphs" employed in (Borgida & Patel-Schneider, 1994), which make the results only hold for the case of same-as restrictions that do not generate "cycles". Furthermore, the algorithm proposed by Frazier and Pitt (1996) does not handle inconsistent concepts, which can easily arise in CLASSIC concepts as a result of conflicts between lower and upper bounds of roles.

Therefore, we have provided an lcs algorithm together with a formal proof of correctness for a sublanguage of CLASSIC with partial attributes, which allows for same-as equalities and inconsistent concepts — the algorithm and proofs can easily be extended to full CLASSIC (Küsters & Borgida, 1999). In this case, the lcs always exists, and it can be computed in time polynomial in the size of the two initial concept descriptions. As shown by Cohen et al. (1992), there are sequences of concept descriptions for which the lcs may grow exponentially in the size of the sequence.

To complete the picture, and as the main part of the paper, we then examined the question of computing lcs in the case of total attributes. Surprisingly, the situation here is very different from the partial attribute case (unlike with subsumption). First, for the language $\mathcal{S}$ the lcs may not even exist. (The existence of the lcs mentioned by Cohen et al. (1992) is due to an inadvertent switch to partial semantics for attributes.) Nevertheless, the existence of the lcs of two concept descriptions can be decided in polynomial time. But if the lcs exists, it may grow exponentially in the size of the concept descriptions, and hence the computation of the lcs may take time exponential in the size of the two given concept descriptions.

As an aside, we note that it has been pointed out by Cohen et al. (1992) that concept descriptions in $\mathcal{S}$ correspond to a finitely generated right congruence. Furthermore, in this context the lcs of two concept descriptions is the intersection of right congruences. Thus, the results presented in this paper also show that the intersection of finitely generated right congruences is not always a finitely generated right congruence, and that there is a polynomial algorithm for deciding this question. Finally, if the intersection can be finitely generated, then the generating system may be exponential and can be computed with an exponential time algorithm in the size of the generating systems of the given right congruences.

The results in this paper therefore lay out the scope of the effect of making attributes be total or partial functions in a description logic that supports the same-as constructor. Moreover, we correct some problems and extend results in the previous literature.

We believe that the disparity between the results in the two cases should serve as a warning to other researchers in knowledge representation and reasoning, concerning the importance of explicitly considering the difference between total and partial attributes.

## Acknowledgments

The authors wish to thank the anonymous reviewers for their helpful comments. This research was supported in part by NSF Grant IRI-9619979. It was carried out while the first author was at the Rutgers University and the RWTH Aachen.